\documentclass[10pt,twocolumn,letterpaper]{article}

\usepackage{iccv}
\usepackage{times}
\usepackage{epsfig}
\usepackage{graphicx}
\usepackage{amsmath}
\usepackage{amssymb}
\usepackage{caption}
\usepackage{multirow,booktabs,colortbl,tabularx}
\usepackage{tabularx}
\usepackage{xcolor}
\usepackage{multirow}
\usepackage{lipsum,mdframed}
\usepackage[export]{adjustbox}
\usepackage{wrapfig,booktabs}

\newcolumntype{C}[1]{>{\centering\let\newline\\\arraybackslash\hspace{0pt}}m{#1}}

\usepackage[pagebackref=true,breaklinks=true,letterpaper=true,colorlinks,bookmarks=false]{hyperref}

\iccvfinalcopy % *** Uncomment this line for the final submission

\def\iccvPaperID{1783} % *** Enter the ICCV Paper ID here

\ificcvfinal\fi

\begin{document}

%%%%%%%%% TITLE
% \title{Single Stage Weakly Supervised Semantic Segmentation from Point Supervision}
\title{Towards Single Stage Weakly Supervised Semantic Segmentation}

\author{Peri Akiva\\
Rutgers University\\
{\tt\small peri.akiva@rutgers.edu}
\and
Kristin Dana\\
Rutgers University\\
{\tt\small kristin.dana@rutgers.edu}
}

\maketitle
\ificcvfinal\thispagestyle{empty}\fi
% \vspace{-15.05em}
%%%%%%%%% ABSTRACT
\begin{abstract}
    \vspace{-0.9em}
   The costly process of obtaining semantic segmentation labels has driven research towards weakly supervised semantic segmentation (WSSS) methods, using only image-level, point, or box labels. The lack of dense scene representation requires methods to increase complexity to obtain additional semantic information about the scene, often done through multiple stages of training and refinement. Current state-of-the-art (SOTA) models leverage image-level labels to produce class activation maps (CAMs) which go through multiple stages of refinement before they are thresholded to make pseudo-masks for supervision. The \textit{multi-stage} approach is computationally expensive, and dependency on image-level labels for CAMs generation lacks generalizability to more complex scenes. In contrary, our method offers a \textit{single-stage} approach generalizable to arbitrary dataset, that is trainable from scratch, without any dependency on pre-trained backbones, classification, or separate refinement tasks. We utilize point annotations to generate reliable, on-the-fly pseudo-masks through refined and filtered features. While our method requires point annotations that are only slightly more expensive than image-level annotations, we are to demonstrate SOTA performance on benchmark datasets (PascalVOC 2012), as well as significantly outperform other SOTA WSSS methods on recent real-world datasets (CRAID, CityPersons, IAD). 
\end{abstract}

%%%%%%%%% BODY TEXT
\vspace{-1.9em}
\section{Introduction}
\vspace{-0.5em}
% \begin{figure}[t]
%     \centering
%     \includegraphics[width=\columnwidth]{images/general/teaser.png}
%     \caption{Generating reliable pseudo-masks even when training from scratch. When training from scratch, our method produces superior pseudo-masks, which leads to better performance. }
%     \label{fig:teaser}
% \end{figure}

The fundamental computer vision task of semantic segmentation seeks to assign class labels to specific pixels in a given input image. The rapid development of deep learning methods has resulted in significant progress in performance \cite{zhao2017pyramid, chen2017deeplab, zhang2018context, long2015fully}, stability \cite{ronneberger2015u, wu2019wider}, and accessibility \cite{NEURIPS2019_9015, tensorflow2015-whitepaper} of semantic segmentation algorithms, often seen in real world applications such as autonomous vehicles \cite{Zhou_2020_CVPR, Kumar_2021_WACV, Varghese_2020_CVPR_Workshops, Blum_2019_ICCV_Workshops, Biasetton_2019_CVPR_Workshops, Meyer_2019_CVPR_Workshops}, precision agriculture \cite{akiva2021aionthebog, akiva2020finding}, medical diagnosis \cite{Yu_2020_CVPR, Wang_2020_CVPR, Lee_2020_CVPR, Ribeiro_2020_CVPR_Workshops, Smith_2019_ICCV_Workshops}, image restoration and editing \cite{Myers-Dean_2020_CVPR_Workshops, Liba_2020_CVPR_Workshops}, sports \cite{Cioppa_2019_CVPR_Workshops}, and remote sensing \cite{akiva2021h2o, Leotta_2019_CVPR_Workshops}. While such algorithms provide insightful information about the scene, it requires large amounts of pixel-wise labeled data \cite{zhou2017scene, lin2014microsoft, Everingham10}, which is often expensive and time consuming to collect \cite{bearman2016s}. To alleviate this requirement, recent efforts have focused on weakly supervised semantic segmentation (WSSS) using image-level \cite{zhou2018weakly, Wang_2020_CVPR, pinheiro2015image, ahn2018learning, araslanov2020single, fan2020cian}, center point \cite{akiva2020finding, bearman2016s}, scribbles \cite{lin2016scribblesup}, or bounding box \cite{dai2015boxsup} labels. The balance between cost and utilization is essential in determining what kind of annotations are needed. Image-level annotations may be cheap to produce, but require more complex networks not practical for real world applications. On the other hand, pixel-wise annotations may be too expensive and time consuming as an up-front cost. 

% \peri{trim this paragraph}
The type of annotation featured in many WSSS methods is image-level labels, providing the least semantic information 
% \faith{maybe use "but with the cheapest annotations" instead} 
but are cheapest to annotate. Methods that use image-level labels often require computationally expensive methods such as multiple networks, region proposal generation, and refinement steps. These methods are often referred to as \textit{multi-stage} weakly supervised semantic segmentation, since they include multiple stages of training and evaluation before performing final inference. Current SOTA WSSS methods utilize class activation maps (CAMs) \cite{zhou2016learning, Oquab_2015_CVPR, selvaraju2017grad} to obtain pixel-wise coverage and localization of objects in the scene. CAMs are often noisy, and only cover the most discriminating parts of objects, making them poor pseudo-masks candidates. For that reason, WSSS methods often resort to multiple stages of refinement before obtaining final pseudo-masks.
% CAM is an attention mechanism 
% typically constitutes the initial stage, in which image level labels are used to train a classification network, from which CAMs are extracted and thresholded to be used as masks for a separate semantic segmentation network. 
% CAMs are often noisy, and only cover the most discriminating parts of the object, making them incomplete and/or imprecise mask candidates. 
% 
% This effect is magnified when CAMs are generated for novel images (i.e. images in the test set but not in the training set).
% \peri{This effect is magnified when class activation maps are extracted for unseen images.
% }
% For that reason, most CAM driven WSSS methods generate pseudo-masks by training and predicting on the same training set, to then train a separate semantic segmentation network.
% to get a final prediction on unseen images. - might be redundant of the previous paragraph
% 
% \peri{Some methods \cite{} generate multiple CAMs as proposal regions and either select or combine them to form the final pseudo-mask. Such operations often add an additional stage to the entire pipeline.}
% 
% Recent work that aims at single stage weakly supervised semantic segmentation \cite{araslanov2020single} through pseudo ground truth generation has shown comparable results to two stage methods.
% 
% 
Such \textit{multi-stage} requirements make adapting those methods to new datasets more difficult. Any change in data distribution requires significant effort, and approaches such as online learning \cite{bottou1998online} become impractical to adapt. 
In this work, we aim to semantically segment images using point annotations (22.1 sec/image in annotation time compared to 20 sec/image for image-level annotations \cite{bearman2016s}) in a \textit{single-stage} approach. 
% 
% Current state-of-the-art (SOTA) in WSSS utilize class activation maps (CAM) \cite{Oquab_2015_CVPR, selvaraju2017grad} in order to localize objects in the scene.
% Such work typically use image level labels to train a classification network, and then extract class activation maps from the last layer to be used as mask predictions of the network. Those predictions are then used as pseudo ground truth for a separate, semantic segmentation network. Such activation maps are often noisy, and only cover the most discriminating parts of the object, making them incomplete and/or imprecise mask candidates. 
% This effect is magnified when class activation maps are extracted for unseen images. For that reason, most class activation map driven segmentation networks train and predict on the same dataset, creating pseudo-masks for that training set, which is then used to train a separate, semantic segmentation network to get a final prediction on unseen images. 
% 

\begin{figure*}[t!]
\setlength\tabcolsep{1pt}
\def\arraystretch{0.5}
\centering
\begin{tabular}{@{}ccc@{\hspace{0.6em}}cccc@{}}
% \begin{tabularx}{\textwidth}{ccccccc}
    % Order: Input Image & Ground Truth  & Ours & Input Image & Ground Truth  & Ours
    \includegraphics[width=0.14\linewidth]{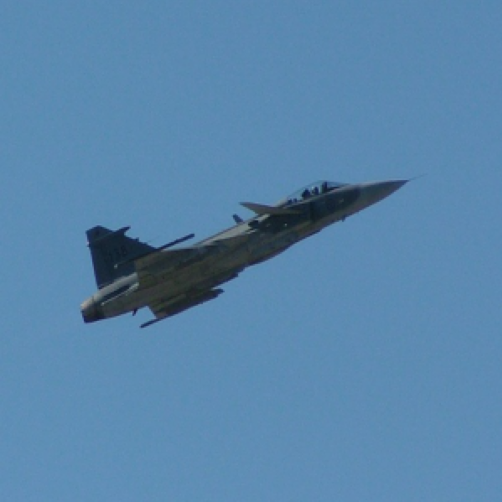} &
    \includegraphics[width=0.14\linewidth]{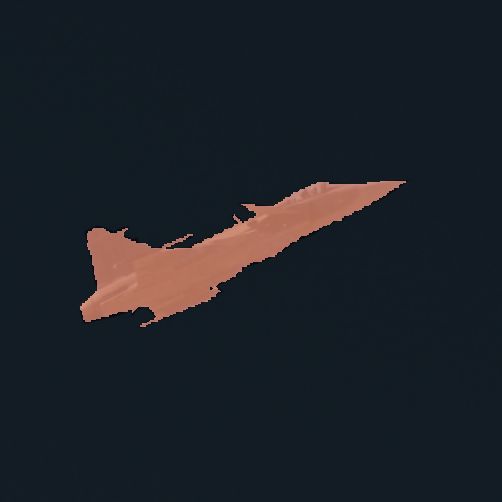} &
    \includegraphics[width=0.14\linewidth]{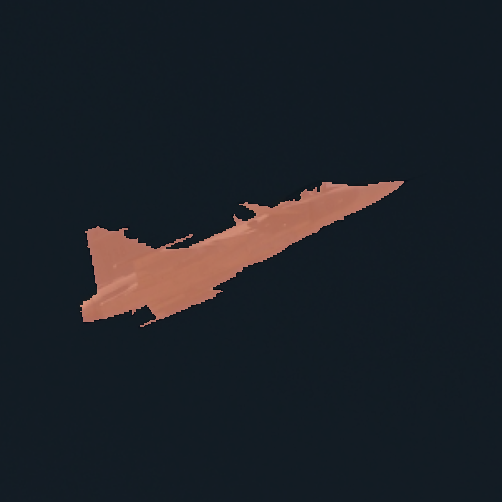} &
    \includegraphics[width=0.14\linewidth, height=0.14\linewidth]{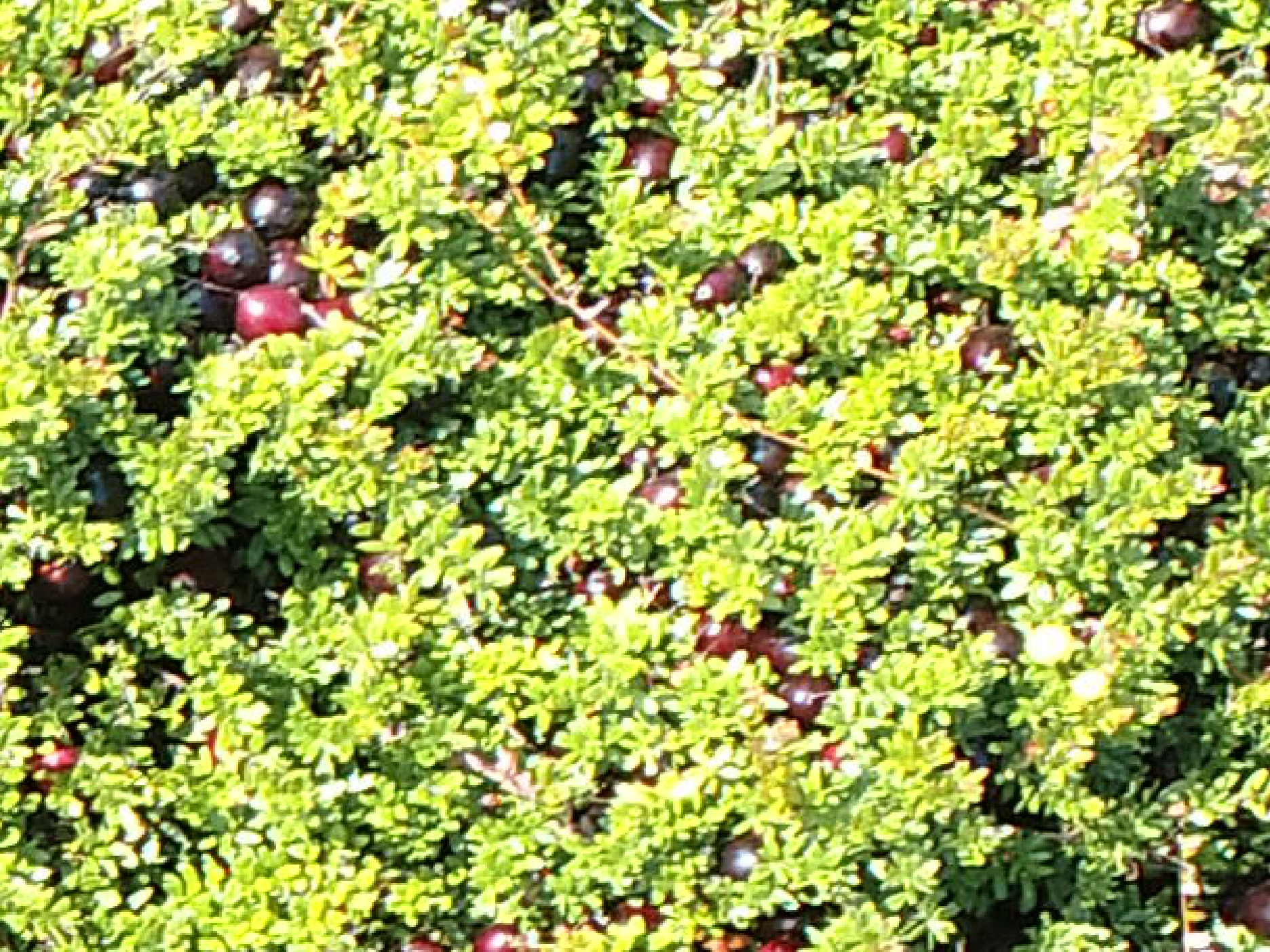} &
    \includegraphics[width=0.14\linewidth, height=0.14\linewidth]{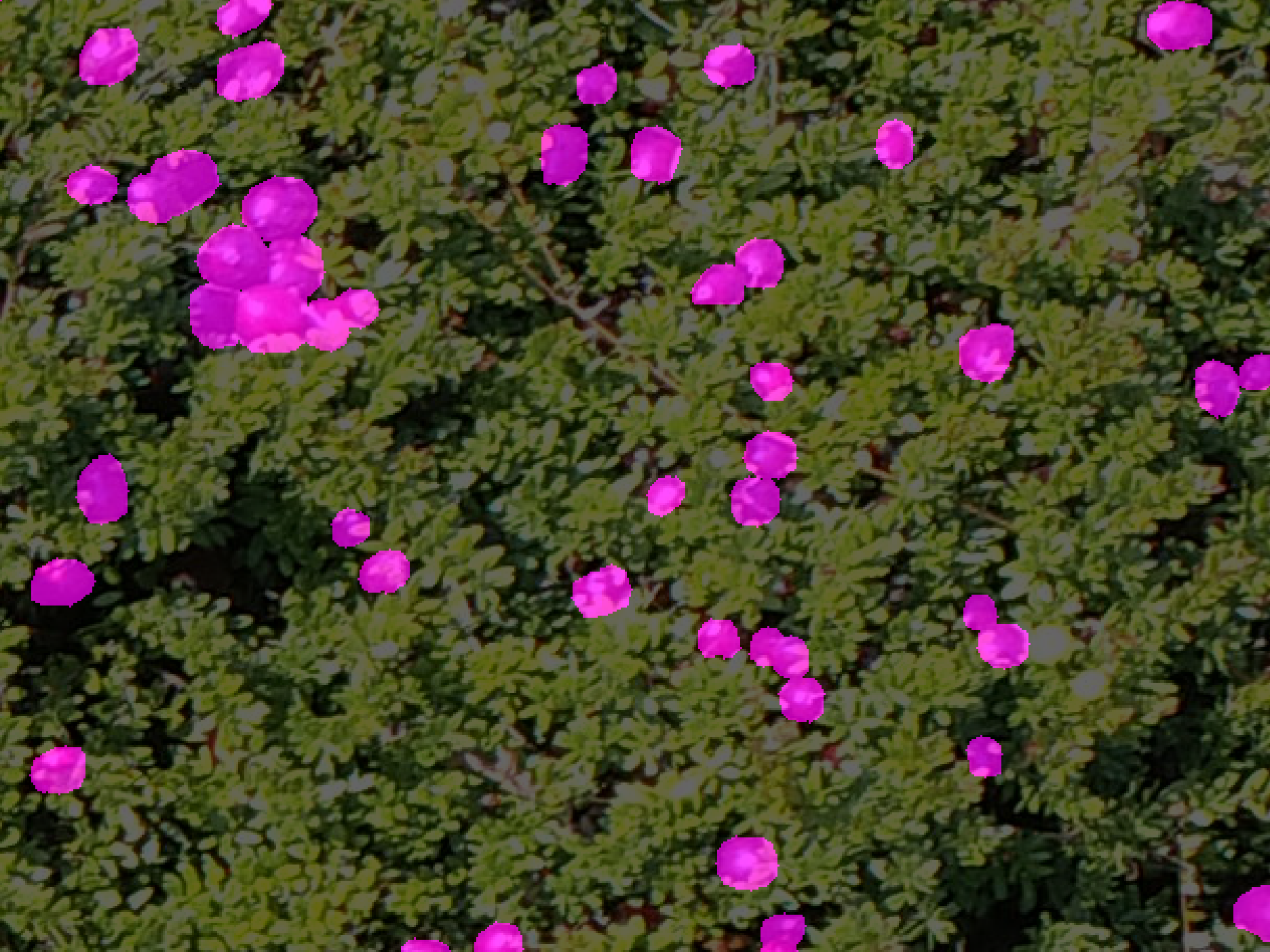} &
    \includegraphics[width=0.14\linewidth, height=0.14\linewidth]{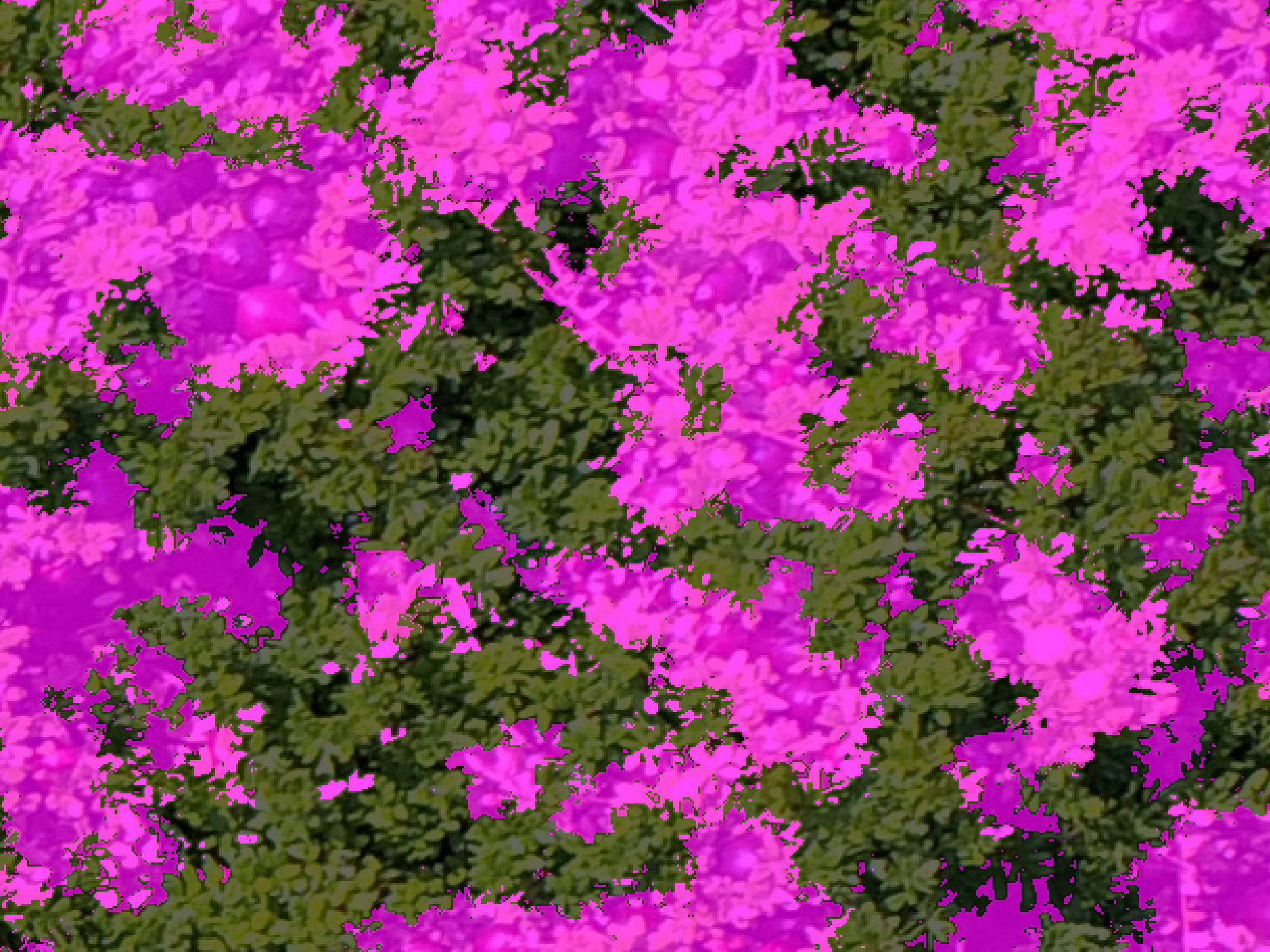} &
    \includegraphics[width=0.14\linewidth, height=0.14\linewidth]{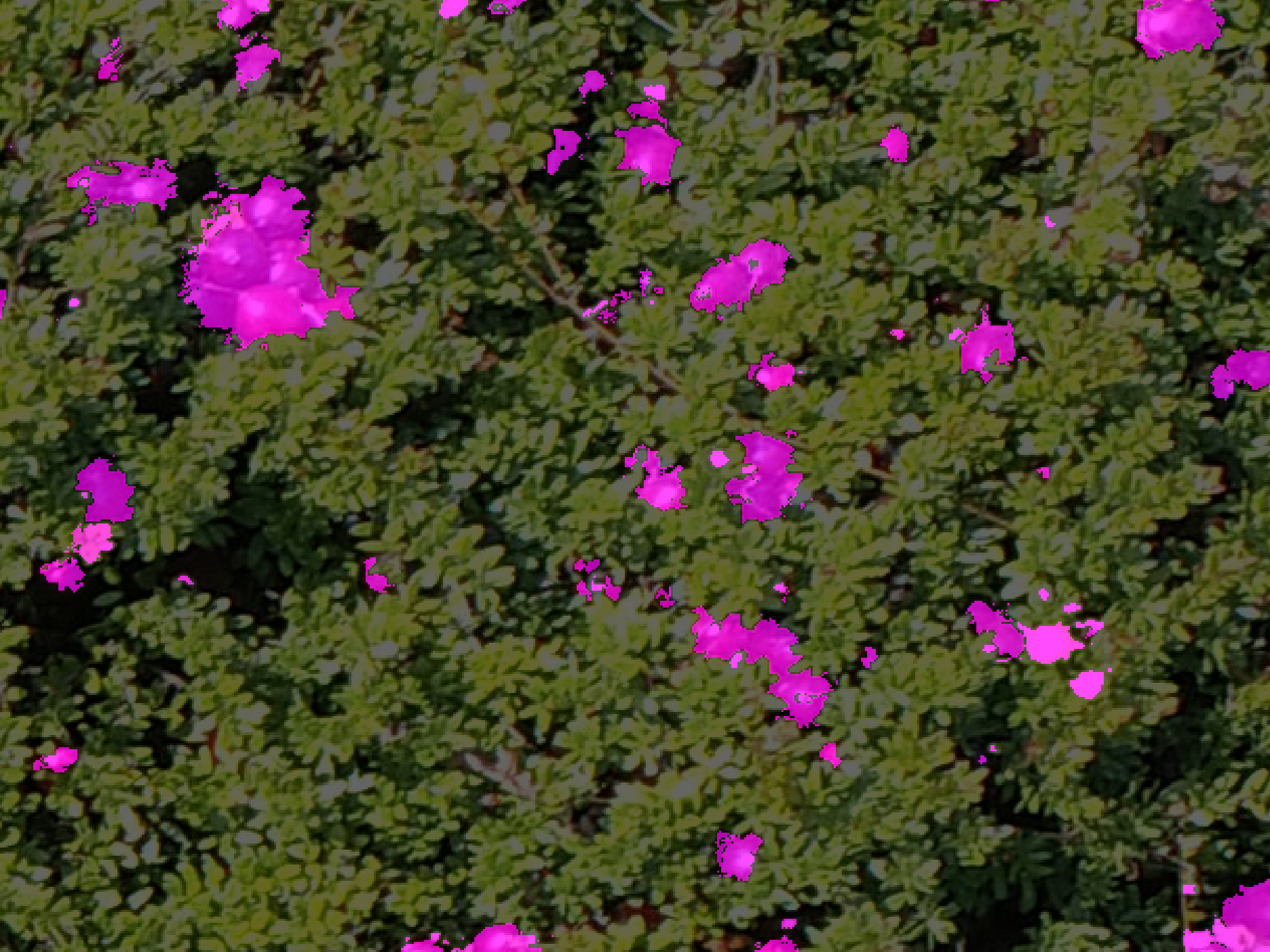} \\
    \includegraphics[width=0.14\linewidth]{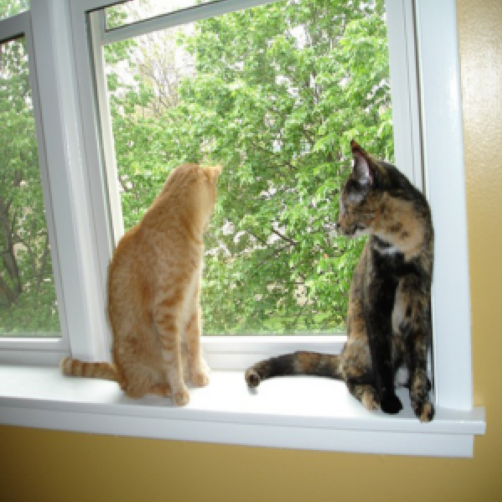} &
    \includegraphics[width=0.14\linewidth]{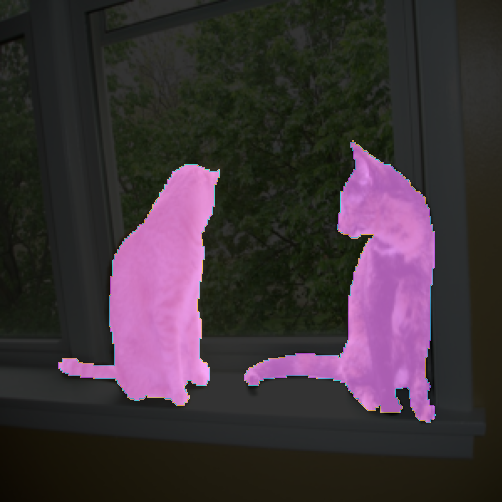} &
    \includegraphics[width=0.14\linewidth]{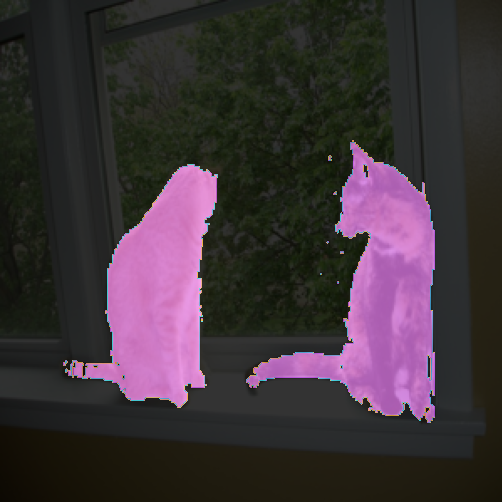} &
    \includegraphics[width=0.14\linewidth, height=0.14\linewidth]{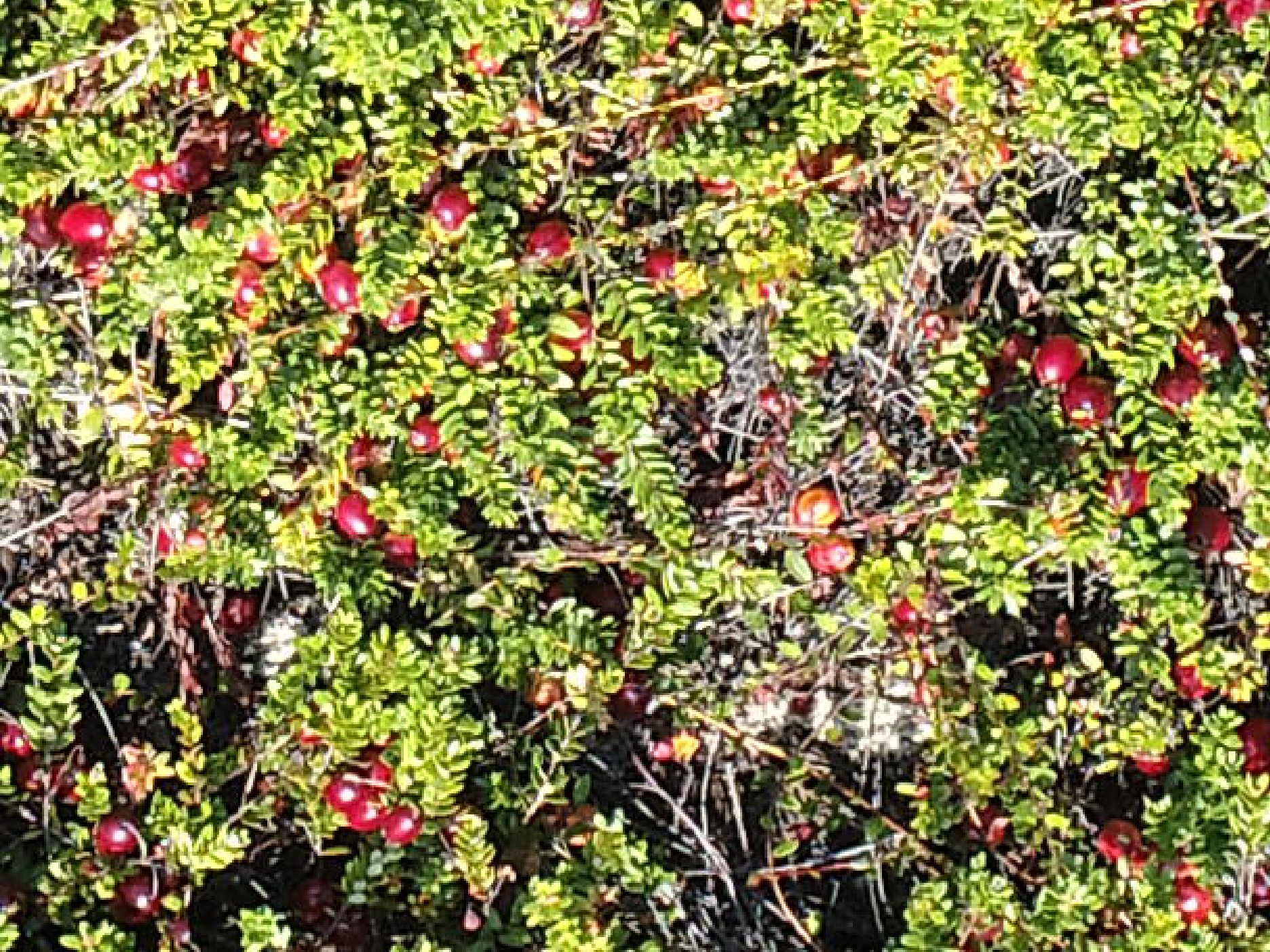} &
    \includegraphics[width=0.14\linewidth, height=0.14\linewidth]{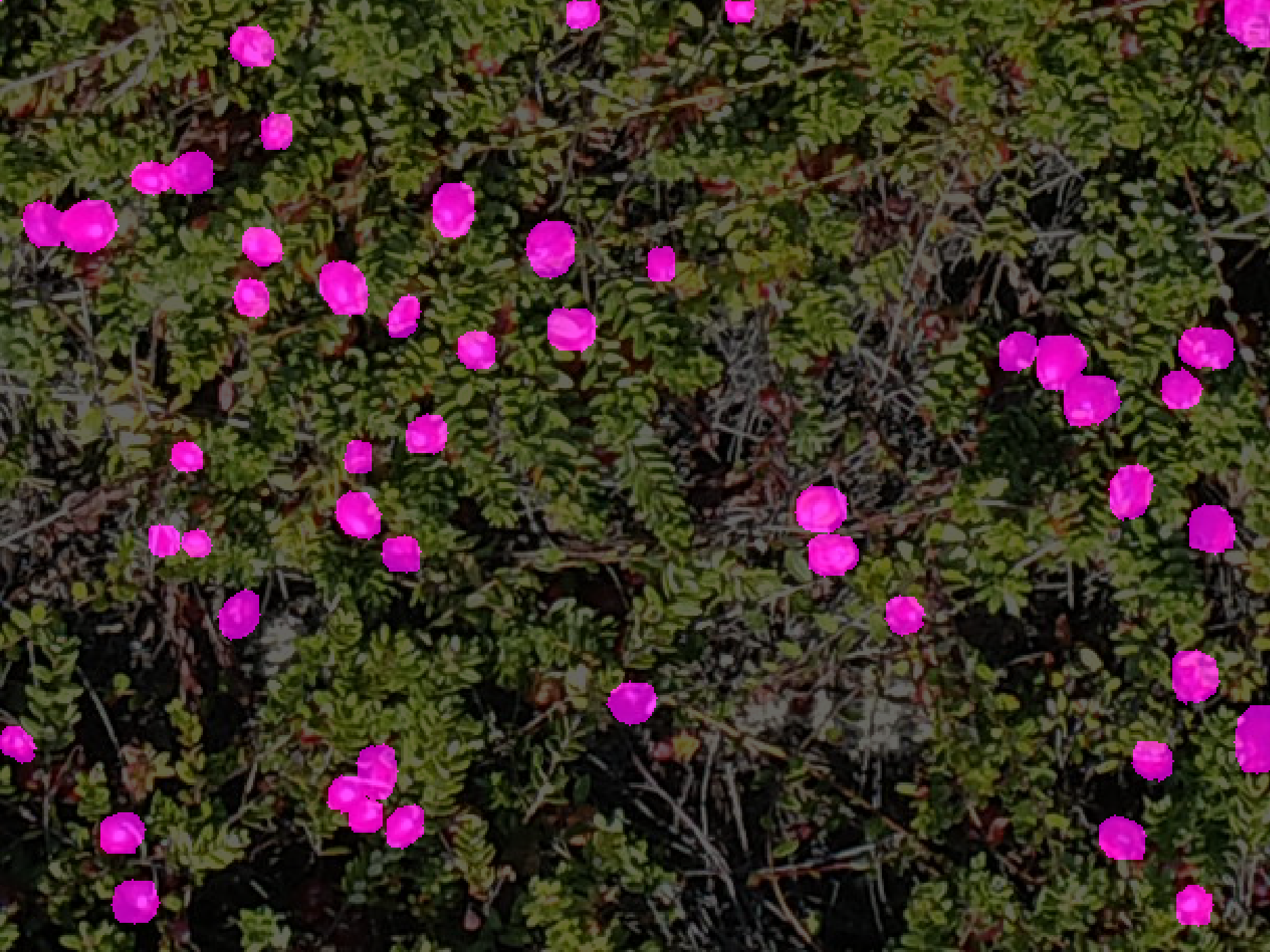} &
     \includegraphics[width=0.14\linewidth, height=0.14\linewidth]{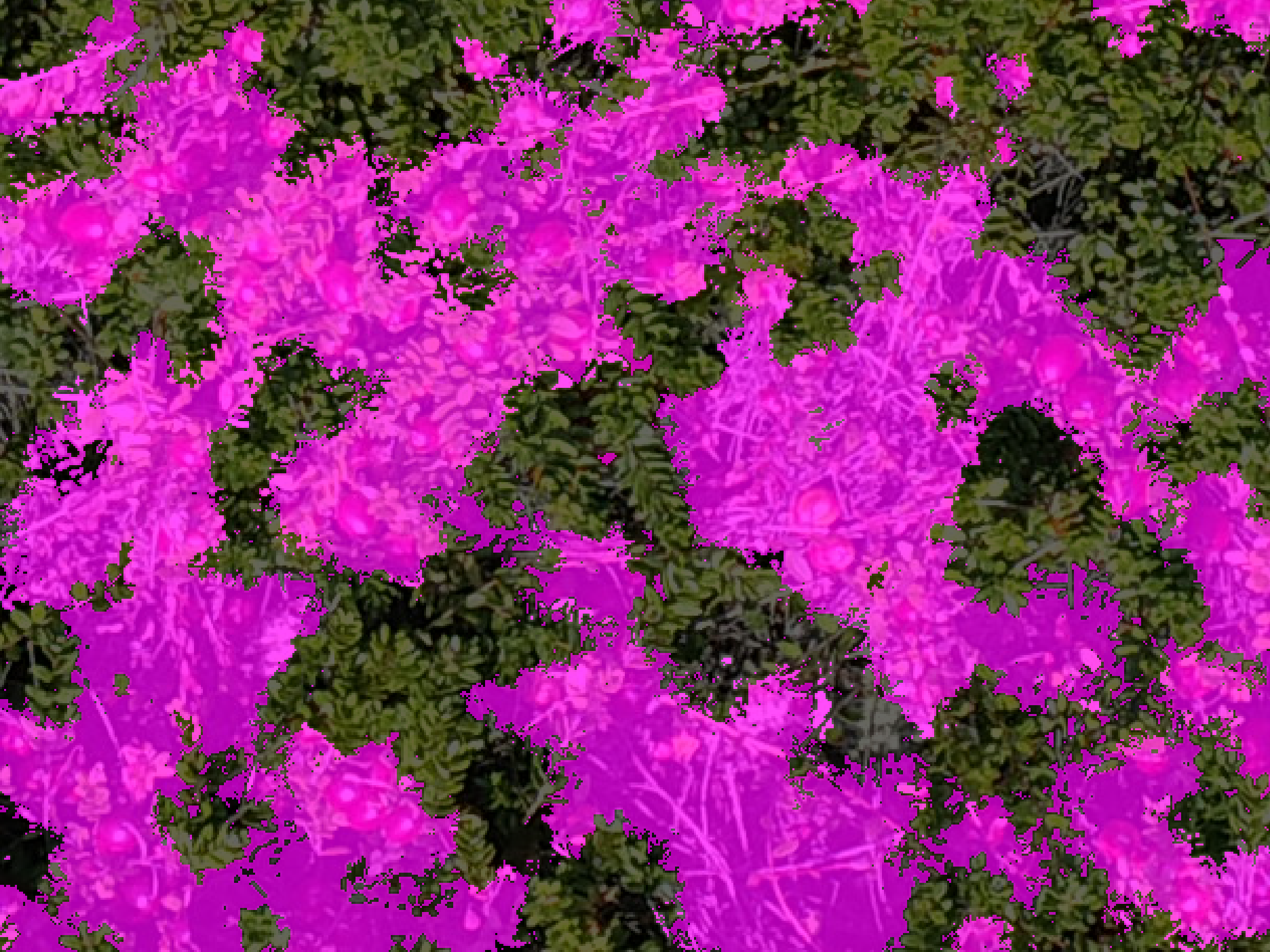} &
    \includegraphics[width=0.14\linewidth, height=0.14\linewidth]{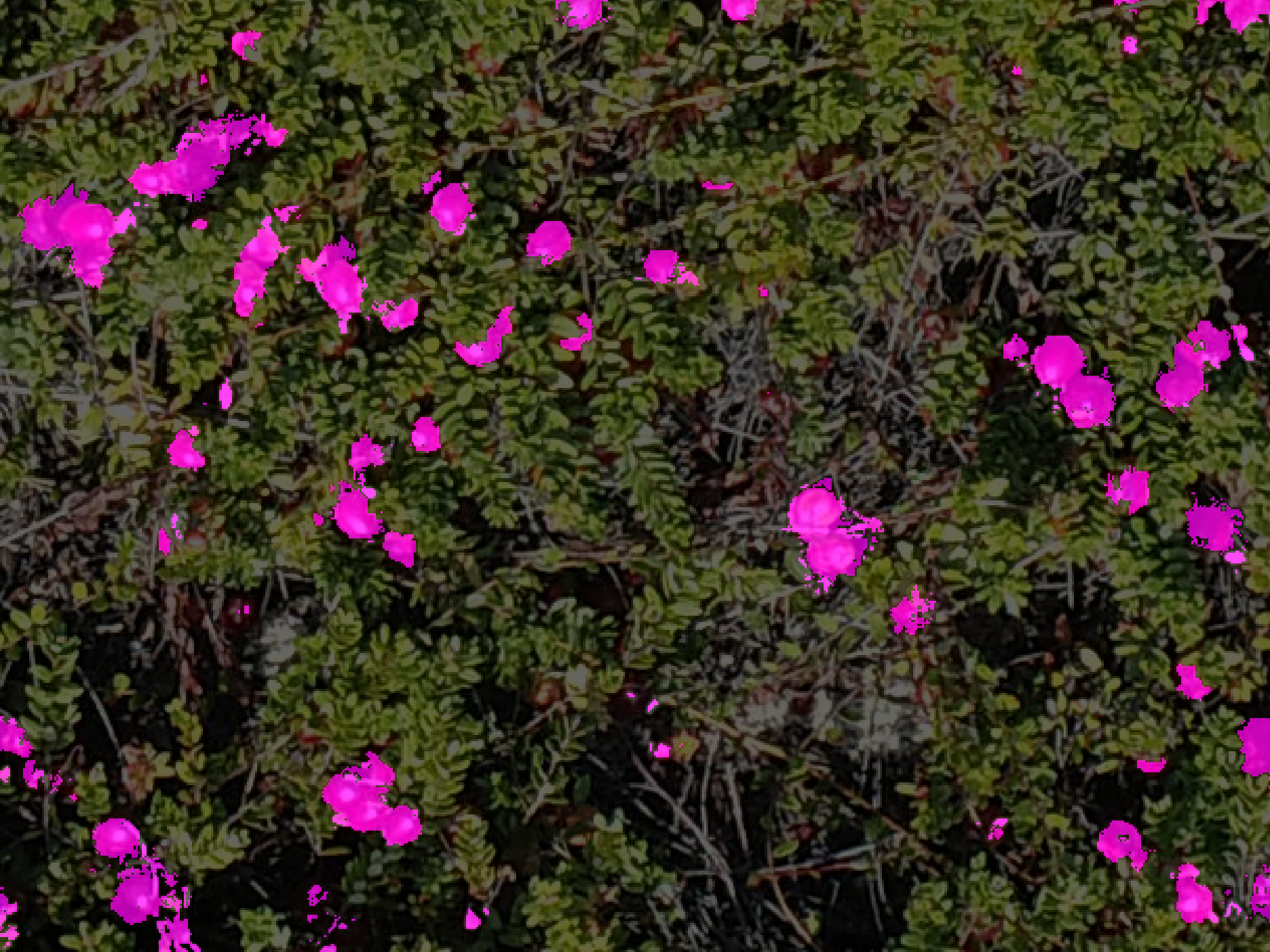} \\
    \includegraphics[width=0.14\linewidth]{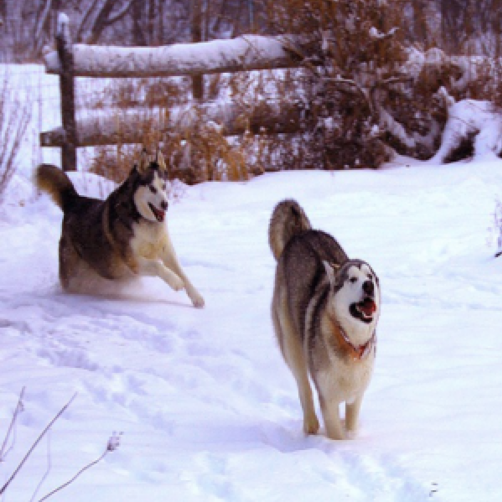} &
    \includegraphics[width=0.14\linewidth]{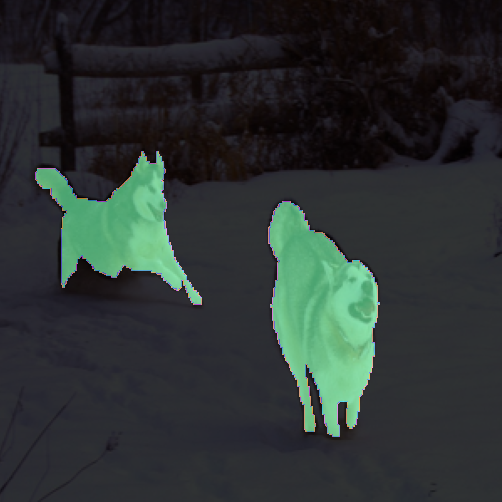} &
    \includegraphics[width=0.14\linewidth]{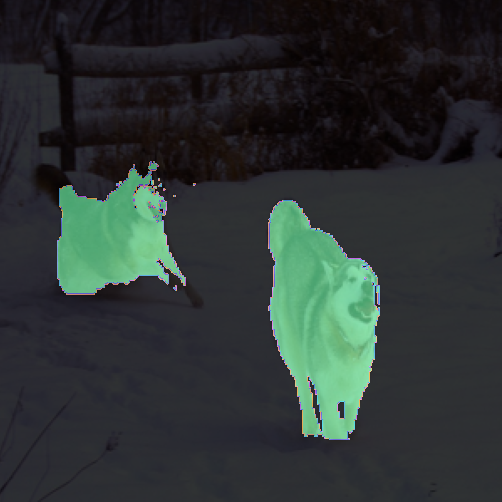} &
    \includegraphics[width=0.14\linewidth]{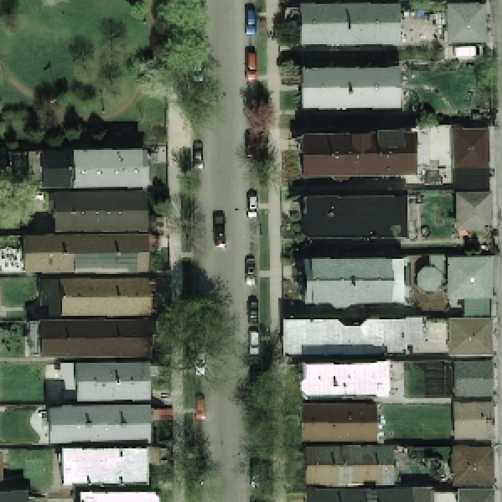} &
    \includegraphics[width=0.14\linewidth]{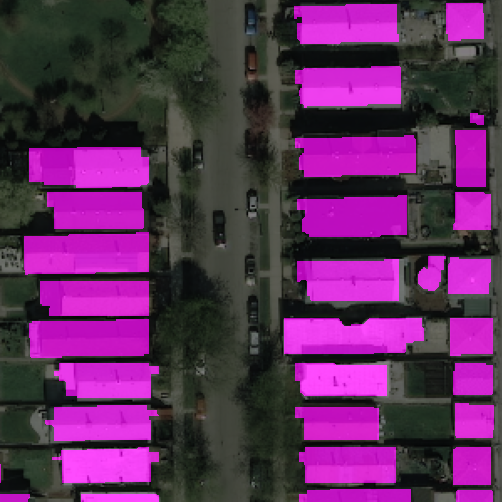} &  
    \includegraphics[width=0.14\linewidth]{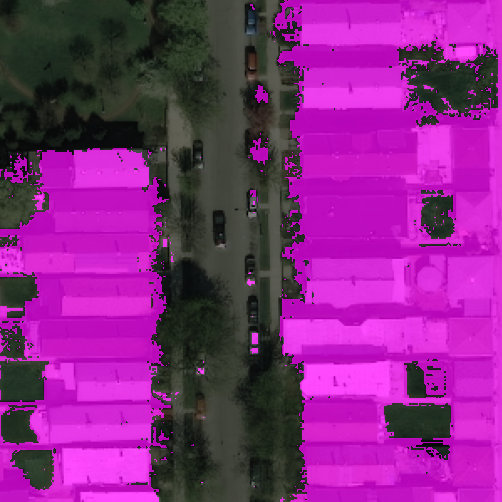} &
    \includegraphics[width=0.14\linewidth]{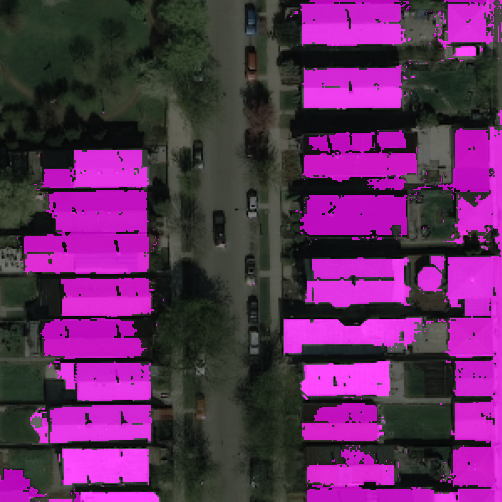} \\
    \includegraphics[width=0.14\linewidth]{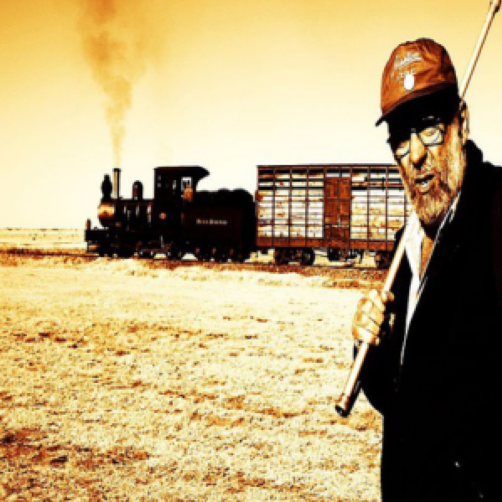} &
    \includegraphics[width=0.14\linewidth]{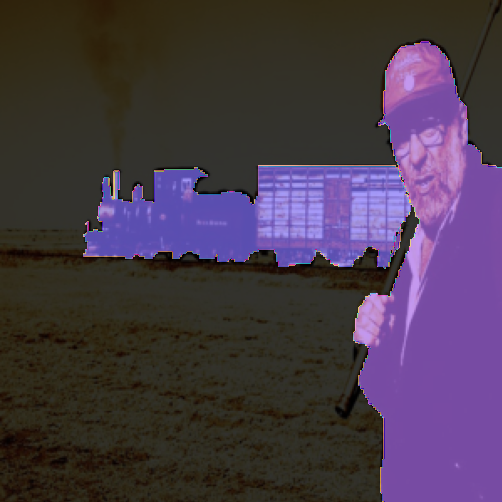} &
    \includegraphics[width=0.14\linewidth]{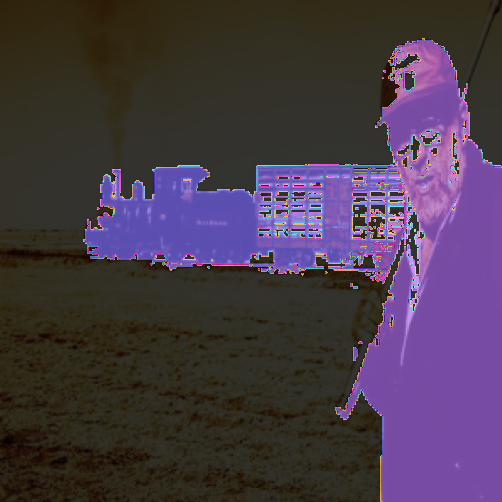} &
    \includegraphics[width=0.14\linewidth]{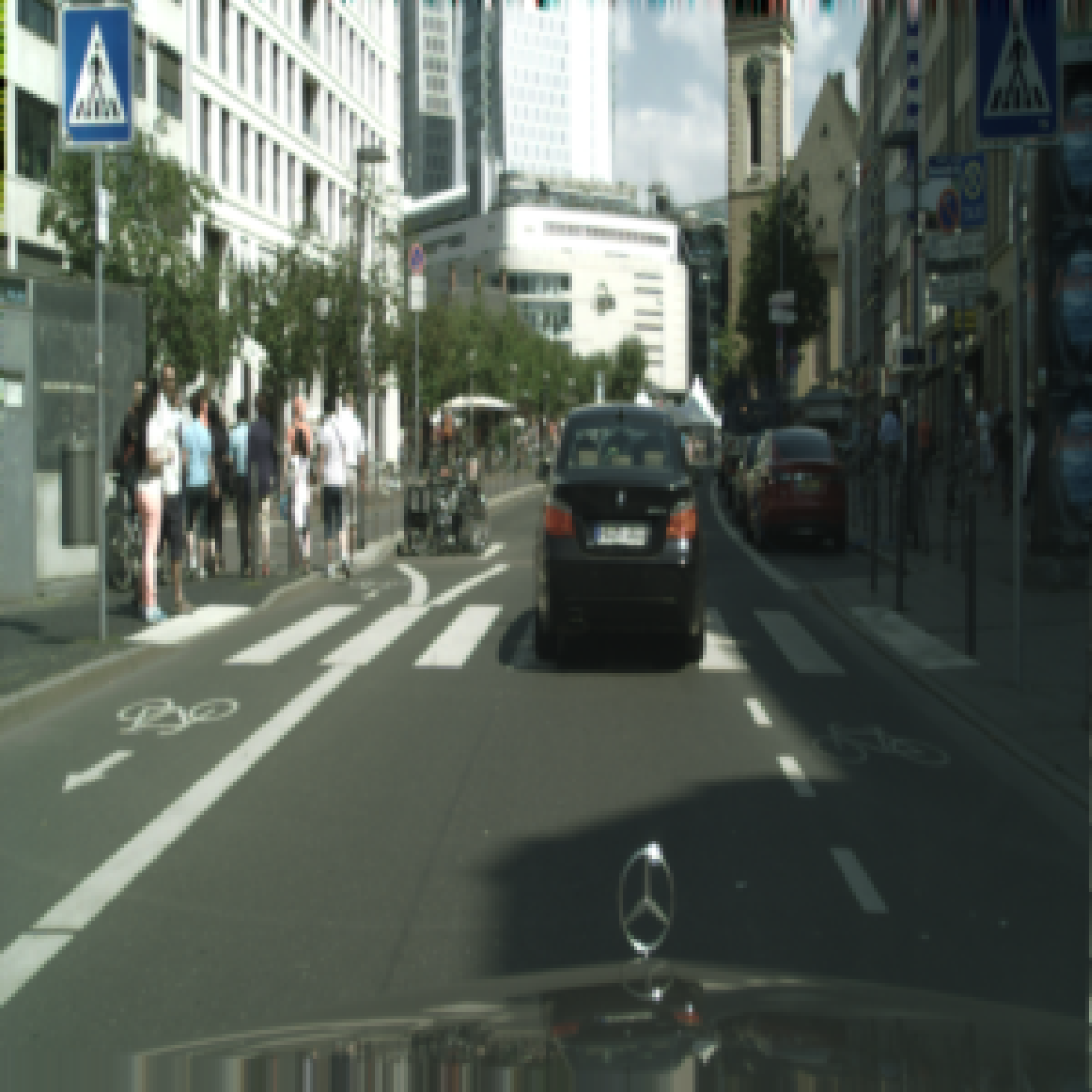} &
    \includegraphics[width=0.14\linewidth]{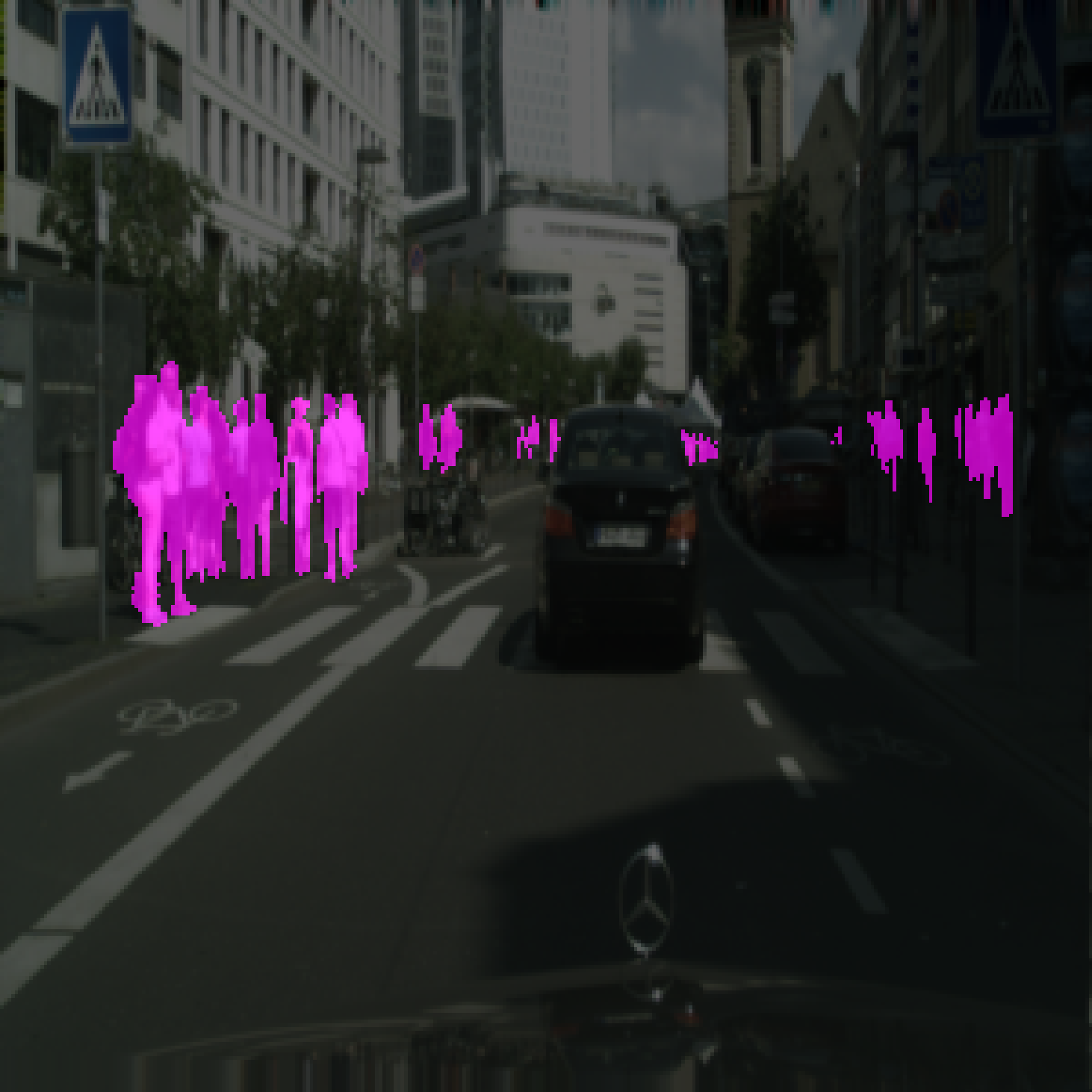} &
    \includegraphics[width=0.14\linewidth]{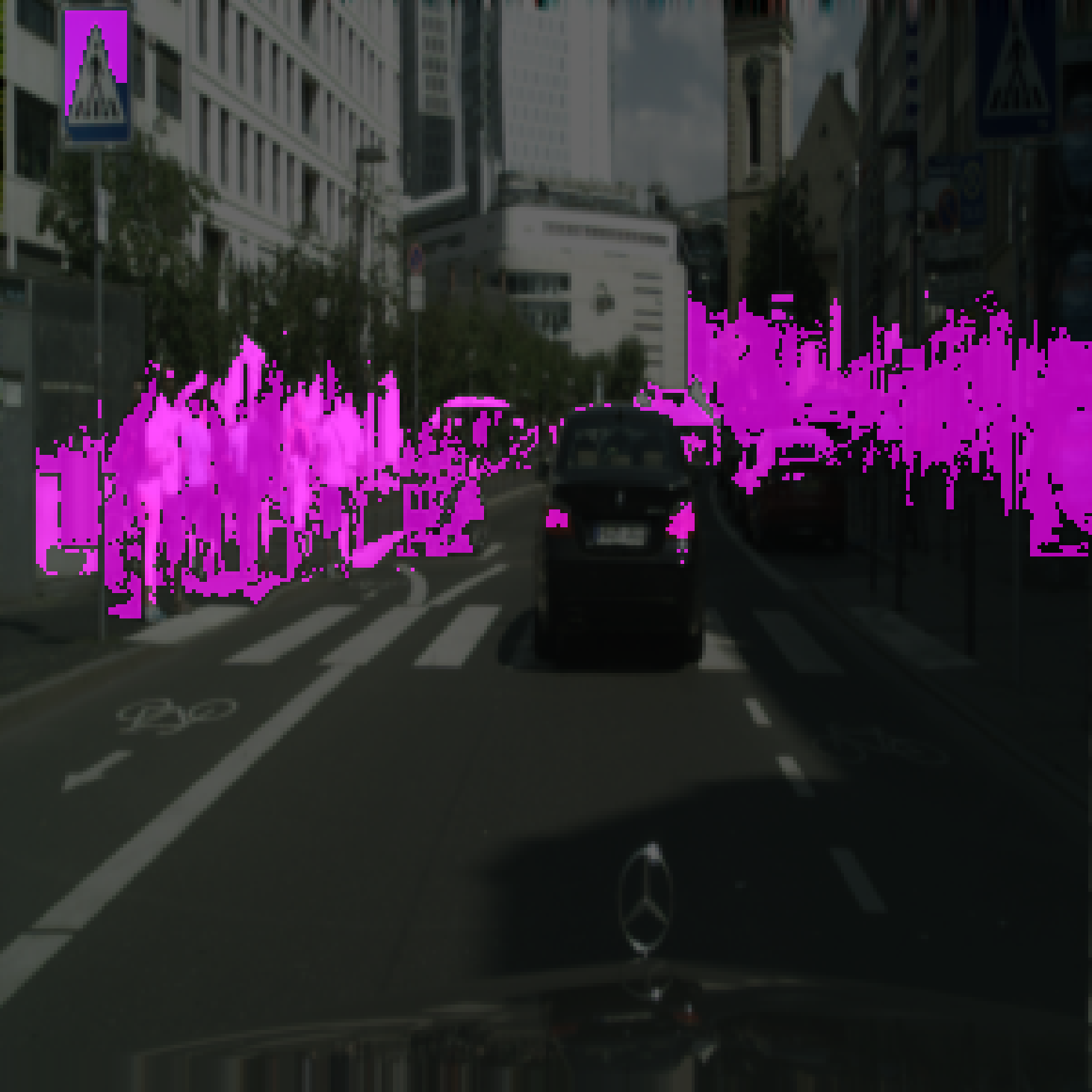} &
    \includegraphics[width=0.14\linewidth]{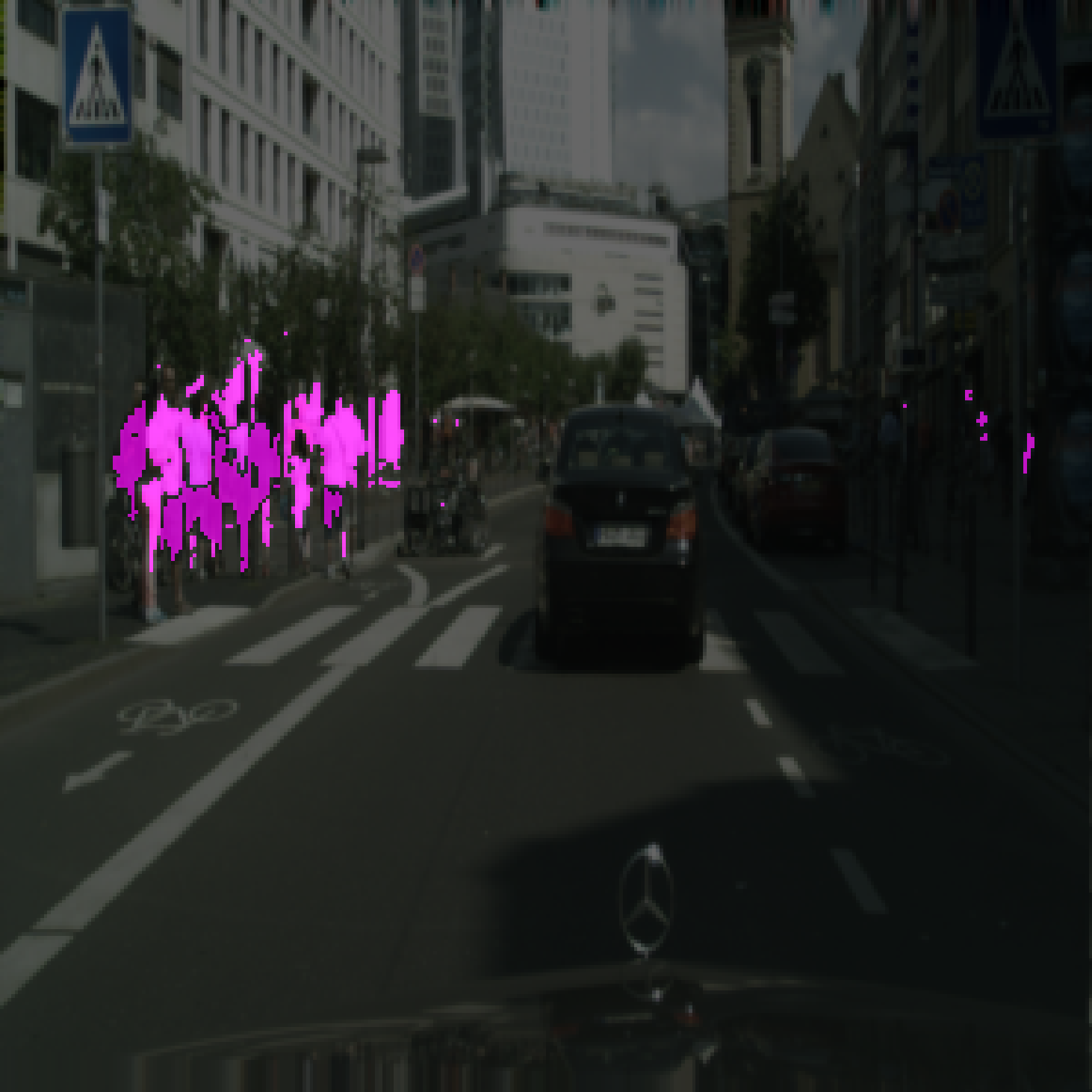} \\
    Input Image & 
    Ground Truth  & 
    Ours &
    Input Image & 
    Ground Truth  & 
    \cite{araslanov2020single} &
    Ours 
\end{tabular}
% \end{tabularx}
\vspace{-0.9em}
\caption{\textbf{Qualitative results} of our method on Pascal VOC 2012 \cite{Everingham10} (left), CRAID \cite{akiva2020finding} (right, first two rows), IAD \cite{maggiori2017can} (right, third row), and CityPersons \cite{Shanshan2017CVPR, Cordts2016Cityscapes} (right, last row) trained with points. Observe that our method provides significantly more refined predictions than our single-stage baseline (trained on image-level) on real world datasets (right side). Best viewed in color and zoomed. Dark gray pixels represent background class.}
\vspace{-0.6cm}
\label{fig:qualitative_results}
\end{figure*}

Recent work has achieved notable improvement at single-stage WSSS \cite{araslanov2020single} through pseudo-mask generation and refinement to obtain comparable results with two stage methods. While resulting performance improvement is significant, the method still lacks generalizability to non-benchmark datasets due to its dependency on a pre-trained backbone. By extension, this dependency limits applicability to real world problems. Pre-trained backbones (trained on the benchmark dataset) are essentially trained classification networks, similar to what is used in multi-stage methods. They provide superior localization and coverage, ``skipping" the challenging stage of generating pseudo-masks in early iterations. From our experiments, competing methods don't generate reliable pseudo-masks without using a pre-trained network, which fundamentally loses its single stage approach when non-standard datasets are used.
% 
% \faith{can probably leave this to related work} 
% 
Performance on standard benchmark datasets is essential to determine the efficacy of novel methods. However, real world applications often necessitate segmenting scenes where image-level labels are not sufficient, e.g. pedestrian, agricultural crops, and biological cells. 
When images have few (or binary) object labels, classification becomes easy, the resulting coarse CAMs (from image-level labels) are enough for classification, but are insufficient for segmentation and therefore produce poor pseudo-masks. This means that dependency on a pre-trained backbone, or any image-level pre-training procedure, can be detrimental for real world applications, which we demonstrate in the experiments (table \ref{tab:craid_results}), results (Figure \ref{fig:qualitative_results}), and supplementary material sections. 

\vspace{-0.4em}
Here, we propose a {\it single-stage} WSSS method that generates reliable pseudo-masks from point annotations. We choose to use point annotations since, while only costing an additional 2 seconds per image in annotation time (20.0 sec/image compared to 22.1 sec/image on average) \cite{bearman2016s}, it provides spatial information essential for correctly localizing and segmenting objects. The method comprises two main novel contributions. First, a point generator component transforms few points to many points using a basic intuition:  
%We craft our framework with the following intuition: 
Given a user-defined object point, the task of sampling {\it another} object point is not so hard. In fact, classical work on random walks in image segmentation can be re-formulated for this problem. Our approach is a point augmentation by iteratively scattering the original points by small affine perturbations followed by random walks.  The point-set obtained by this iterative scatter-then-walk procedure is termed {\it point blot}, analogous to ink blot.
This point blot generation stage is entirely deterministic and does not require any training.  Furthermore, the resulting point blot has significantly more utility compared to the original point-clicks.
% 
% We combine local and global contextual cues obtained by our Relative Entropy Perturbation Expansion (REPE) and Growing Attentional Distance Maps Aggregation (GADMA) algorithms to obtain on-the-fly, reliable ground truth masks to train a weakly supervised semantic segmentation network. 
% REPE makes use of an initial set of points to generate a second set of points that is consistent with the first. Intuitively, we divide-and-conquer the overall local expansion into smaller sub-problems, which are easier to accomplish than full segmentation. 
The second contribution  
% \faith{also re the last point; maybe contribution instead of intuition?} 
in our framework is the {\it expanding distance fields}, a new instantiation of the classic distance fields \cite{breu1995linear} that is refined within the learning pipeline to capture changing inter-class distances.
% 
% can be considered a point blot to pseudo-mask transformation which uses spatial inter-class distance transforms (called Expanding Distance Field) and feature refinement to generate reliable pseudo-masks for supervision. 
When considering early training iterations of an un-trained network, outputs are expected to be noisy and unstable, producing unreliable pseudo-masks. To mitigate such errors, we employ pixel adaptive convolution refinement network along with our expanding distance fields. The former rectifies locally inconsistent regions conditioned to local statistical representation, and the latter filters wrongly activated regions, and stabilizes training by preventing accumulation of bias in generated pseudo-masks. 
Our main contributions is a a method to generate pseudo-masks from points that use  point blot generation and expanding distance fields. Notably, our method eliminates the dependency on pre-trained backbones and classification tasks for WSSS.  We achieve SOTA performance on single-stage WSSS. We achieve significantly better performance in non-benchmark datasets from important application domains.

\vspace{-0.8em}
\section{Related Work}
\vspace{-0.4em}
\subsection{Semantic Segmentation}
\vspace{-0.4em}
Semantic segmentation is a dense image prediction task that predicts class labels for every pixel in a given image. The quick progress in deep learning and convolutional neural networks \cite{krizhevsky2012imagenet, simonyan2014very, szegedy2015going, he2016deep} has powered the development of the fully convolutional network (FCN) \cite{long2015fully}, which is the basis to many current SOTA semantic segmentation methods \cite{zhao2017pyramid, chen2017deeplab, zhang2018context, ronneberger2015u}. Typical design of semantic segmentation networks utilize encoder-decoder architectures, in which deep features are learned, and up-sampled to match the input image size. More recent work improve this base design by incorporating skip connections \cite{ronneberger2015u}, contextual information \cite{zhang2018context}, self-attention mechanisms \cite{fu2019dual}, enlarged receptive fields \cite{zhao2017pyramid, he2015spatial}, pyramid pooling \cite{zhao2017pyramid, chen2017deeplab, he2015spatial, lin2017feature}, and refiner networks \cite{Lin_2017_CVPR}. U-Net \cite{ronneberger2015u} adopts skip connections to fuse information from low level layers into high level layers. PSPNet \cite{zhao2017pyramid} features a multi-scale pyramid pooling module applied to
% 
% \faith{"that applies on" sounds clunky. Maybe "applied to" instead?}
% 
sub-regions that act as the contextual representation of that region. A similar type of pyramidal pooling scheme is employed in DeepLap \cite{chen2017deeplab} and Feature Pyramid Networks (FPN) \cite{lin2017feature}. EncNet \cite{zhang2018context} encodes multi-scale contextual features using a Gaussian kernel. With similar goals in mind, ASPP \cite{he2015spatial} captures additional contextual information through dilated convolution layers, which enlarge the receptive field of the layer without added parameters. 
While those networks often provide SOTA performance, they still require expensive computational cost and fully supervised ground truth. 

% Current state-of-the-art utilize encoder-decoder networks largely inspired 
\vspace{-0.4em}
\subsection{Class Activation Maps and Region Proposals}
\vspace{-0.3em}
The activated neurons of a deep learning network with response to an input image are called class activation maps (also referred to as CAMs or attention maps) \cite{zhou2016learning}. They represent regions the network finds most distinctive for a given class label. Initial work leveraging CAMs 
% 
% \faith{which approach? you haven't laid it out in text yet.(besides the section title)} 
% 
were used for object localization \cite{Oquab_2015_CVPR, zhou2016learning, tang2016large, bilen2016weakly, li2016weakly} and network interpretability \cite{selvaraju2017grad}, but were recently adopted for semantic and instance segmentation tasks \cite{fu2019dual,ahn2018learning,Wang_2020_CVPR, fan2020cian, chaudhry2017discovering, zhou2018weakly}. Most approaches utilize CAMs as masks, region proposals, or auxiliary data used to generate labels for segmentation methods. Since CAMs tend to be noisy and irregular in shape, much focus in the WSSS domain has been devoted towards refining outputs to improve CAM coverage accuracy and consistency. SEAM \cite{wang2020self} addresses the volatility of CAM outputs by enforcing consistency between transformed inputs using a siamese network. Instead of learning better CAMs, SSDD \cite{shimoda2019self} resorts to two post-processing refinement networks (also read as two additional stages) to improve outputs before generating pseudo-masks.
% SSDD \cite{shimoda2019self} employs a four-stage method in which, after a classifier is trained and CAMs are generated, the output pseudo-masks are further refined with two cascaded networks.
\vspace{-0.15em}
\subsection{Weakly Supervised Semantic Segmentation}
\vspace{-0.35em}
The majority of work done in the WSSS domain is accomplished in a multi-step process: train a classification or segmentation network, apply the network on the training set to extract CAMs, which are then refined and thresholded before used to train a separate segmentation network. 
% Such pseudo-masks are often generated using a combination of activation maps extracted from the trained network as an indication of discriminative parts of the object. 
Early work like BoxSup \cite{dai2015boxsup} utilize bounding boxes to update pre-defined region proposals to generate ground truth masks for the training set. AffinityNet \cite{ahn2018learning} leverages image level labels to generate affinity labels obtained through selection of high confidence points on amplified CAMs. 
Similarly, PRM (Peak Response Map) \cite{zhou2018weakly} back-propagates through local extrema points in attention maps to generate instance-wise pseudo-masks. 
Additional methods \cite{Wang_2020_CVPR, fan2020cian, chaudhry2017discovering, huang2018weakly, wang2020self, lee2019ficklenet} follow a similar approach using image level labels for pseudo-mask generation.
While those methods achieve significant performance improvement over previous weakly supervised methods, they all utilize ground truth labels during inference time to eliminate activation maps of classes not present in the image. That practice is consistent with the \textit{multi-stage} process of acquiring pseudo-masks for the training set, during which labels are assumed to be present. 

Recent single stage WSSS methods \cite{akiva2020finding, araslanov2020single, huang2018weakly, hong2016learning, hong2017weakly, papandreou2015weakly, pinheiro2015image, roy2017combining} are less common due to the challenge of implicitly obtaining reliable spatial and contextual information from weak labels. Triple-S \cite{akiva2020finding} uses point supervision and shape priors as spatial and contextual cues for the network. However, the use of shape priors is highly restrictive, and explicitly provides spatial and contextual information to the network, making the method too task specific. In contrast, Araslanov \etal \cite{araslanov2020single} train a segmentation-aware classification network using normalized global weighted pooling (nGWP), iterative mask refinement, and focal mask penalty. Normalized global weighted pooling allows concurrent classification and segmentation training, while the output mask prediction is iteratively refined using Pixel Adaptive Convolution (PAC) layers introduced in \cite{su2019pixel}. 
While \cite{araslanov2020single} shows significant improvement in single stage WSSS, the method requires a pre-trained backbone to achieve good performance. The use of pre-trained weights removes biases and randomness present during initial training steps, allowing for superior pseudo-mask generation. Generally, as seen in \cite{Oquab_2015_CVPR, selvaraju2017grad}, a trained classification network provides ``free" localization of the objects by locating peaks in class activation maps. Such localization would not be available unless the backbone is pre-trained, or trained first. Since \cite{araslanov2020single} requires training a classification network as an additional stage, then it may be considered a two-stage approach when used for non-benchmark datasets. 
% \peri{maybe add about how classification task on few or binary class tasks hinders the performance of the segmentation since many times the features are not refined/localized well. Especially in sparse/high count scenes such as CRAID}
% If we consider the training of the classification network as an additional stage, then \cite{araslanov2020single} may be considered a two-stage approach when used for non-benchmark datasets. 
% 
% \peri{Additionally, other two stage methods already train a classification task, which is what the backbone is pre-trained for. }
% 
% \peri{From our experiments, training \cite{araslanov2020single} from scratch has degraded performance and stability significantly compared to other multi-stage methods. [- this line is also used in the intro and results sections. is it important here?]}
% 
% 
% 
Similar to \cite{araslanov2020single}, we also adopt our own version of Pixel Adaptive Convolution layers \cite{su2019pixel} for feature refinement, and subsequent pseudo-mask generation.
% 
% Contrary to that approach, our method aims to leverage activation maps in an end-to-end manner ...

\begin{figure*}[t!]
    \centering
    \includegraphics[width=0.99\textwidth]{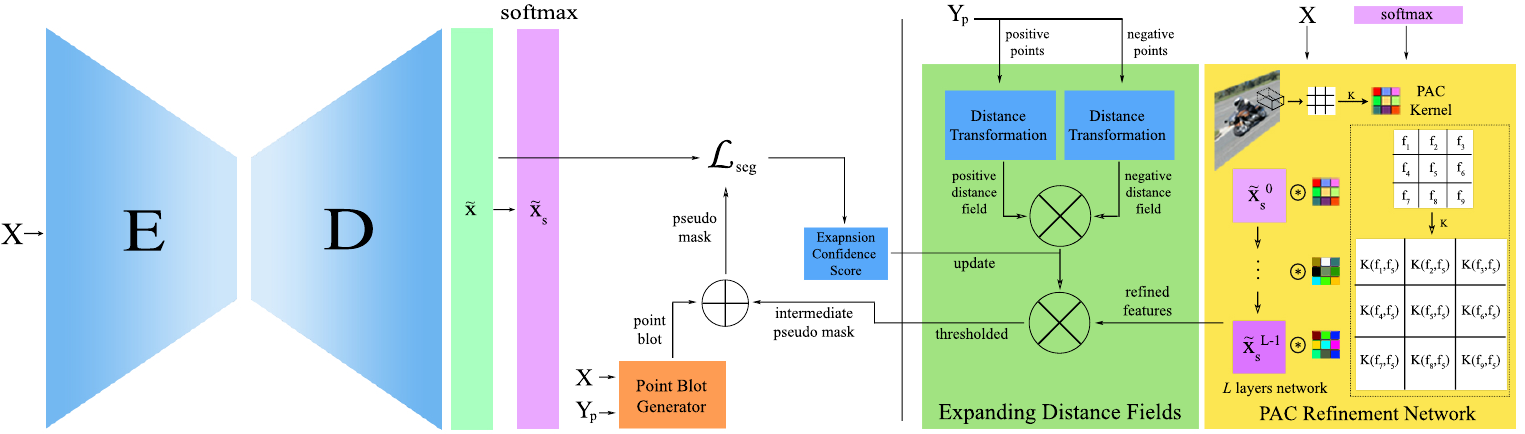}
    \vspace{-0.9em}
    \caption{\textbf{Pseudo-Mask from Points (PMP)} overall architecture. Input image is fed to a fully convolutional network, and supervised by a pseudo-mask generated by the Expanding Distance Fields (\textit{green box}) and Point Blot Generation (\textit{orange box}) modules. The network's output features is refined with accordance to local statistics of underlying features and color distributions using our Pixel Adaptive Convolution Refinement Network variant (\textit{yellow box}). The refined features are then multiplied element-wise with expanded distance fields to generate intermediate pseudo-masks. When training a network from scratch, early iterations tend to be unstable, producing noisy outputs and therefore unreliable pseudo-masks. Our novel Expanding Distance Fields allow from-scratch training by preventing accumulation of error in generated pseudo-masks. Best viewed in color and zoomed.}
    \label{fig:architecture}
    \vspace{-2.2em}
\end{figure*}

\vspace{-0.9em}
\section{Pseudo-Masks from Points}
\vspace{-0.4em}
% 
% 
% \subsection{Problem Formulation}
% Our proposed method is split into two parts: global and local pseudo mask generation. 
The motivation behind
% \faith{motivation behind? or you could mirror the intuition thing from the intro with something like The main inspiration behind our method?} 
our method is to obtain reliable, on-the-fly pseudo-masks from initial points to train a semantic segmentation network. Intuitively, the better the ground truth labels, the better the network's performance. Pseudo-masks are typically obtained through some thresholding of high confidence (high activation) features. When training from scratch, such features tend to be noisy, and directly thresholding such features will generate poor pseudo-masks which will result in sub-optimal training and performance. We address that challenge by using the Expanding Distance Fields module (section \ref{sec:expanding_distance_fields}), which filters wrongly activated regions, and captures and amplifies correctly activated regions. It also introduces a new aggregation approach and expansion mechanism 
% \faith{is the aggregation approach the expansion mechanism or is that separate?}
that alleviates overfitting to features around ground truth points. We also employ a Point Blot Generator (section \ref{sec:repe}) and its point blot output to provide superior utility compared to points alone, capturing additional locally available contextual information, and accelerating training progress. As seen in Figure \ref{fig:architecture}, we incorporate a feature refinement network (section \ref{sec:pac_network}) to work in tandem with our Expanding Distance Fields to produce intermediate pseudo-masks, which are superimposed with  point blots to make the final pseudo-masks for supervision.

% In order for the method to be able to generate reliable pseudo-masks when trained from scratch, localizing objects is essential to guide pseudo-masks into high confidence regions. Our method follows three main steps to achieve
% 
% This method preserves consistency both globally and locally using GADMA and REPE modules. 
% Expanding Distance Fields module utilizes our PAC refinement network (section \ref{sec:pac_network}), distance map aggregation (\ref{sec:aggregation_dm}), and expanding confidence mechanism (\ref{sec:growing_dm}) to obtain \peri{intermediate pseudo-masks}, which are superimposed with point blots generated by the Point Blot Generator (section \ref{sec:repe}) to make the final pseudo-mask for supervision.  
% 
% elements to obtain a final pseudo ground truth label: 1) global contextual \peri{attention} consistency, and 2) local \peri{uniformity - wrong word}. Global contextual \peri{attention} consistency (\peri{change this name}) is obtained through our classification-aware segmentation approach and attention distance map aggregation algorithm detailed in sec \ref{sec:global}. Local \peri{uniformity - wrong word} is accomplished through our relative entropy perturbation expansion explained in section \ref{sec:repe}. Once both, local and global masks are obtained, (\peri{wordy?}) we simply combine them to make the final pseudo ground truth mask. 
% 
% \vspace{-0.2em}
\subsection{\normalsize \textbf{Pixel Adaptive Convolution Refinement Network}}
\vspace{-0.4em}
\label{sec:global}
\label{sec:pac_network}
% \peri{Attention Guided Global Mask?}
% 
% 
Our implementation of Pixel Adaptive Convolution Refinement Network (also referred to as PAC Refinement Network or PAC Refiner)
% \faith{what's the first task?} 
stems from pixel adaptive convolution layers \cite{su2019pixel}. 
% 
% \faith{you either need this (<-) sentence or the very first sentence and not both} 
% 
They allow for dynamic modification of kernel weights based on some underlying conditions, and are
% 
% \faith{this sentence should come after the next one} 
% 
commonly used in feature refinement work \cite{wannenwetsch2020probabilistic, kumar2021syndistnet, dundar2020panoptic, xu2020learning, tanujaya2020semantic, araslanov2020single} with modified kernel functions. 
% suitable for the task in hand. 
% 
% \faith{this (<-) sentence probably belongs in related work??}  
% 
Here, we use PAC layers to dampen activated regions in the output features that are not locally consistent.
% that are not consistent with the local color distribution of the input image or deemed locally inconsistent with themselves. 
% relevant by local output features.
% 
% \faith{we should workshop this paragraph bc it's not very efficient and could also probably be 2 paragraphs}
% 
% We employ these conditions 
% 
% \faith{ik you mentioned the conditions before but what are they? the color thing?} 
% 
Our PAC Refinement Network considers the local standard deviation of the input image in color space and the local mean in feature space %within the local region of the kernel.
% for the adaptive convolution layer kernel function. 
% 
% The method aims 
to amplify output regions with low standard deviation %in color space 
and high mean. Given network output $\Tilde{x}$ and its corresponding softmax $\Tilde{x}_s$, we feed $\Tilde{x}_s$ to an $L$ layer network consisting of pixel adaptive convolution layers with kernel $k$ as a function of the corresponding pixel values in $X$ and features in $\Tilde{x}^{(l-1)}_s$ (the output of previous layer in the PAC refinement network), where $l$ is the current layer. More specifically,

\vspace{-0.5em}
\begin{equation*}
    \Tilde{x}_s^{l} = \Tilde{x}_s^{(l-1)} \circledast k,
\end{equation*}

\vspace{-0.7em}
\noindent where

\vspace{-0.5em}
\begin{equation*}
    k_{i,j} = - \frac{|X_{i,j} - X_{l,n}|}{\sigma_{i,j}} \mu_{i,j}.
    \vspace{-0.4em}
\end{equation*}
$(i,j)$ correspond to the current location of the kernel, $(l,n)$ are all neighboring pixels of $(i,j)$, $\Tilde{x}_s^{(l-1)}$ is the output of the previous PAC Refiner layer, $\sigma_{i,j}$ is the standard deviation of the current kernel region in $X$, and $\mu_{i,j}$ is the mean value of the current kernel region in $\Tilde{x}^{(l-1)}_s$. Note that $X$ is normalized between 0-1. 

Consider a region over an edge, in which we can expect a larger standard deviation in color intensity, which will result in a smaller kernel weight (dampening kernel). Similarly, the mean over the kernel region in feature space depicts both local similarity and relevancy of those pixels. Low similarity and/or relevancy will reduce the kernel weight over that region. When applied over the entire image, we dampen all locally inconsistent activated regions in $\Tilde{x}_s$.
% , and enforce local consistency to the output. 
The overall operation is structured as an $L$ pixel-adaptive convolution layer network with varying kernel sizes, dilations, and strides. Note that since kernel weights are functions of local color and feature statistical representation (standard deviation and mean), the refined features are the output of a single forward pass of the network. There are no learned weights in this operation, making it computationally inexpensive. 

% We repeat this for \textit{t} iterations in which $\Tilde{x}_s$ is updated. Results of this method in early epochs can be seen in figure \ref{fig:attentional_dm_ag}.
% Note that when the standard deviation
% 
\vspace{-0.3em}
\subsection{Expanding Distance Fields}
% \vspace{-0.3em}
\label{sec:expanding_distance_fields}
Expanding Distance Fields aim to impose global consistency and correct localization in the refined feature space (obtained from \ref{sec:pac_network})
% directed by our novel Expanding Distance Fields.
%
% \faith{you start your sections with sentences like this that make me think the module you refer to at the beginning is different from the EDF that you end the sentence talking about, but based on the section title I think they're the same???} 
%
by leveraging background (negative) and object (positive) point annotations to generate distance fields (section \ref{sec:aggregation_dm}).
% that focalize 
% \faith{I googled this and it's actually not a word this time} 
% refined features towards correct regions. 
These distance fields are then updated by our expansion mechanism (section \ref{sec:growing_dm}) which allows the distance field to incrementally incorporate more refined features into the final output.
% Expanding Spatial Pass Band \faith{!!! \(<3\)}
\vspace{-1.em}
\subsubsection{Distance Field Aggregation}
\vspace{-0.8em}
% \paragraph{Distance Fields Aggregation}
\label{sec:aggregation_dm}
The use of distance fields \cite{breu1995linear} is common in interactive segmentation methods \cite{grady2006random,rother2004grabcut,gulshan2010geodesic,hariharan2011semantic,bai2014error,li2018interactive,xu2016deep,maninis2018deep,jang2019interactive}, where it is used as auxiliary data produced from user inputs such as points and scribbles. Here, we use it as a point-guided filter to enforce object localization consistency in the refined features and subsequent generated pseudo-masks. Our usage of distance fields filters is essential in stabilizing training in early iterations, during which output features lack sound structure to make reliable pseudo-masks.
\begin{figure}[t!]
    \centering
    \includegraphics[width=0.93\columnwidth]{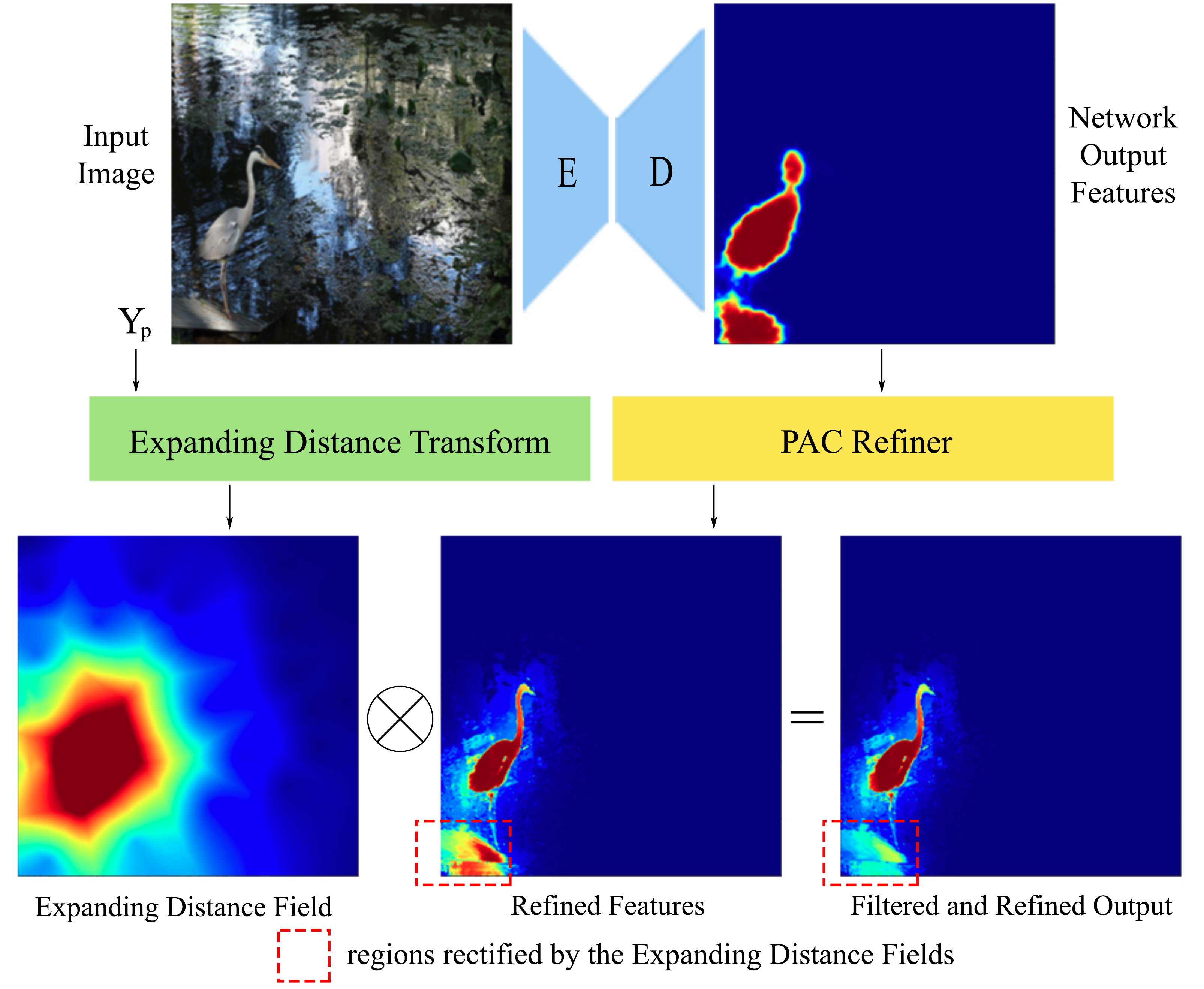}
    \vspace{-1.0em}
    \caption{\textbf{Expanding Distance Fields and PAC Refinement Network} joint utility example on Pascal VOC 2012 \cite{Everingham10}. The network output features are fed to the PAC Refiner, which output is multiplied element-wise with the expanding distance fields obtained from point annotations and expansion confidence score. Observe that the rock is wrongly activated for the bird class, which is dampened by the expanding distance field. The regions highlighted by the red dash boxes indicate wrongly activated regions and their dampened outputs. The final output is thresholded to make the intermediate pseudo-mask. Epoch-by-epoch progress of pseudo-masks, refined features, and expanding distance fields is available in the supplementary material. Best viewed in color and zoomed.}
    \vspace{-1.7em}
    \label{fig:attentional_dm_ag}
\end{figure}
Distance fields are computed by taking the minimum Euclidean distance between a given point and the rest of points present in the scene. Given image $X \in \mathcal{R}^{H \times W \times 3}$ and ground truth points $Y_p \in \mathcal{R}^{H \times W \times 1}$, where $Y_p(i,j) \in \{0,1,\cdots, \mathcal{C}+1\}$ (label $\mathcal{C}+1$ represents background points),
% Note that we also sample background points noted as points with labels $21$ (more details in section \ref{sec:experiments}). 
we use $Y_p$ to obtain class-wise distance fields $\mathcal{D} \in \mathcal{R}^{(C+1) \times H \times W}$, where $C$ is the number of classes. For example, to generate a distance field $\mathcal{D}_{c}$ for some class $c$, we compute the value of  $\mathcal{D}({c,i,j})$ at location $(i,j)$ using $\mathcal{D}(c,i,j|Y_p) = \min \sqrt{(i-p_{c,i})^2 + (j-p_{c,j})^2}$,
% 
% \begin{equation*}
%     \mathcal{D}(c,i,j|Y_p) = \min \sqrt{(x-p_{c,x})^2 + (y-p_{c,y})^2},
% \end{equation*}
% 
where $p_c$ is a point in $Y_p$ that belongs to class $c$ (also referred to as positive points). We repeat this for every point and class in the image to obtain $D \in \mathcal{R}^{(C+1) \times H \times W}$, including for background (or negative) points. Typically, such distance fields are concatenated to the input image of interactive segmentation methods. Instead, we leverage the distance fields to enforce object localization consistency on intermediate pseudo-masks. We invert the normalized background distance field $D_{\mathcal{C}+1}$, and perform element-wise multiplication with all other distance maps: $\mathcal{D}_{c} = (1 - \mathcal{D}_{\mathcal{C}+1}) \odot \mathcal{D}_{c} ~\forall ~c \in \{1,\cdots, \mathcal{C}\}$.
% 
% \begin{equation*}
%     \mathcal{D}_{c} = (1 - \mathcal{D}_{21}) \odot \mathcal{D}_{c} \forall c \in \{1,2,\cdots, 20\}
% \end{equation*}
% 
Inverting $D_{\mathcal{C}+1}$ imposes low values in regions known to belong to the background class. By taking the element-wise product between $D_{\mathcal{C}+1}$ and all other distance maps, we remove regions in $D_{c}$ that may be ambiguous or inconsistent with the underlying object's location. This can also be observed in Figure \ref{fig:attentional_dm_ag}, where the wrongly activated region (marked with a red box) is dampened by the distance field.

\vspace{-1.25em}
\subsubsection{Expansion Mechanism}
\vspace{-0.5em}
% \paragraph{Expansion Mechanism}
\label{sec:growing_dm}
Using points as seeds to create distance fields inherently creates bias towards the regions around those seeds, especially when objects are large. For that reason, we employ our novel expansion mechanism, which aims to represent increasing dependability of the model on the refined features. Typically, outputs tend to be noisy in early training iterations, and as training progresses we expect better output feature representations. If the distance fields are used alone without the expansion mechanism, the intermediate pseudo-masks tend to provide partial coverage for images with large objects. Instead, we define an expansion confidence score, $\mathcal{E}_{score}$, which is a function of the network's learning progress. In initial stages of training, we consider the seed point as the pixel with highest confidence, corresponding to the value of 1. As the network learns features corresponding to that class, we incrementally lower the highest confidence threshold. By doing so, we expand the distance field from the seed point outward, essentially enlarging the region of high confidence
and allowing more refined features in to be included in the output. If done indefinitely, the generated refined features will be ``fully trusted," meaning the entirety of the refined features will be included in the output. Formally,
\vspace{-0.5em}
\begin{equation*}
    \gamma = \frac{\mathcal{L}^{(e-1)}}{\mathcal{L}^{(e)}} - 1,
\end{equation*}
% 
% \vspace{-.5em}
\begin{equation*}
    \mathcal{E}_{score} = \mathcal{E}_{score} + max (min(\gamma, \eta), \omega),
\end{equation*}
where $\mathcal{L}^{(e)}$ is the accumulated loss at epoch $e$, $\gamma$ is the improvement ratio between the current and previous epochs, and $\eta$ and $\omega$ are the upper and lower limits for confidence improvements to be added to $\mathcal{E}_{score}$ at that epoch. Note that performance degradation at a given epoch will result in a lower confidence score for the next epoch. We use the confidence score to modify our distance map aggregation by adding it to the distance fields, and clipping any values below 0 and above 1 as follows,
\begin{equation*}
    \vspace{-0.2em}
    % \mathcal{D} = \mathcal{D} + \mathcal{C}_{score} 
    \mathcal{D}_{c,x,y} = 
    \begin{cases}
      1 & \text{if $\mathcal{D}_{c,x,y} + \mathcal{E}_{score} \geq 1$} \\
      0 & \text{if $\mathcal{D}_{c,x,y} + \mathcal{E}_{score} \leq 0$} \\
      \mathcal{D}_{c,x,y} + \mathcal{E}_{score} & \text{otherwise}\\
    \end{cases}
    \vspace{-0.2em}
\end{equation*}
% \vspace{-0.em}
where $x$ and $y$ represent all possible locations of distance field for class $c$. Classes not present in the image are ignored. Note that we use different expanding confidence scores for the background and the objects, and the object expanding scores increase 2 times faster than the background expanding score. The importance of this module is also visually demonstrated in the supplementary material through epoch-by-epoch distance field instances and their corresponding pseudo-masks. 

% where $i$ is an element in $\mathcal{D}$.
% 
% Let image $X \in \mathcal{R}^{H \times W \times 3}$ and ground truth points $Y_p \in \mathcal{R}^{H \times W \times 1}$ be an input sample 
% 
% obtain regions with distinctive features which correspond to activated regions in the class activation map. We then
% 
% 
% 
\vspace{-1.1em}
% \paragraph{Summarizing paragraph - needs a name}
% \vspace{-.4em}
% \label{sec:attention_dm}
% \paragraph{Summarizing subsection - needs a name}
\paragraph{The final step} \hspace{-1.em} of the Expanding Distance Fields module performs an element-wise product between the refined features and the aggregated distance fields, followed by thresholding, to obtain the final pseudo-mask. 
% The importance of both elements is essential when training the network from scratch. 
Since the PAC refinement network only smooths and ensures local consistency, without global perspective, it often has activated regions that are not part of the objects. By multiplying its output with the aggregated distance fields, we spatially constrain the class activation maps to regions determined to be relevant by the distance fields. This is visually demonstrated in Figure \ref{fig:attentional_dm_ag} which shows the transition between each stage up to the final pseudo-mask.
% This stage is especially important when training a model from scratch since early iterations tend to produce noisy and unreliable outputs. 

\vspace{-.1em}
\subsection{Point Blot Generator}
\vspace{-.1em}
\label{sec:repe}
% \peri{local consistency}

The purpose of this method is to generate a set of new local ground truth pixels from image $X$, and annotated points $Y_p$ through iterative operations of perturbations and random walks over the input image $X$. 
% \faith{put free back in this sentence somewhere}
% 
% \peri{features obtained through a single convolution layer with an affinity kernel \cite{}.} 
The set of new ground truth pixels, named point blots, capture neighboring pixels that are ``obviously" part of the object. Such additional pixels are essential in the early iterations because they provide reliable baseline pseudo-masks before the network is able to generate meaningful features. The role of these point blots decreases as the intermediate pseudo-mask generated by the PAC Refiner and Expanding Distance Fields improve.

\begin{figure}[t!]
    \centering
    \includegraphics[width=0.88\columnwidth]{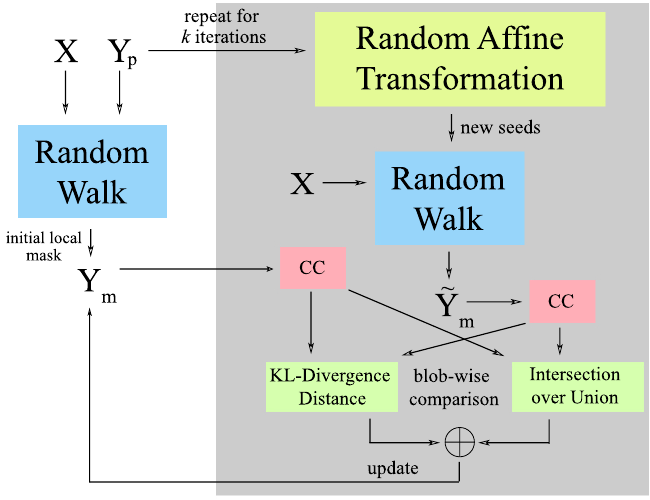}
    \vspace{-0.8em}
    \caption{\textbf{Point Blot Generator} pipeline. The module generates initial point blots using input image and ground truth points. Initial point blots are then iteratively updated conditioned to coverage matching and underlying color distribution similarity of current and candidate blobs. Candidate blobs are generated through perturbations of initial points followed by random walks in color space, which are then separated into candidate blobs using the connected component algorithm \cite{dillencourt1992general} (noted as CC).}
    \vspace{-2.1em}
    \label{fig:REPE}
    
\end{figure}

Let image $X \in \mathcal{R}^{H \times W \times 3}$ and ground truth points $Y_p \in \mathcal{R}^{H \times W \times 1}$ be an input sample to the Point Blot Generation module. We obtain an initial mask, $Y_m$, using a random walk over $X$ with $Y_p$ as seeds. Then, we perturb $Y_p$ using a random affine transformation to obtain new points $\tilde{Y_p}$, which are used as seeds for a random walk over $X$ to generate a candidate mask $\Tilde{Y_m}$. While we can guarantee that all points in $Y_p$ lay on the correct objects, we cannot assume the same for $\Tilde{Y_p}$, and consequently cannot assume that $\Tilde{Y_m}$ is a good candidate mask as a whole. Instead, we partition $Y_m$ and $\Tilde{Y_m}$ into current and candidate blobs, $B,\Tilde{B}$, using the connected component algorithm \cite{dillencourt1992general}, with each current blob $b \in B$ corresponding to a candidate blob $\tilde{b} \in \tilde{B}$. We then calculate the Kullback–Leibler divergence (KLD) distance \cite{kullback1951information} between the distributions of the underlying image features enclosed by the pixels of $b$ and $\tilde{b}$. A candidate blob is accepted as an expansion to its corresponding current blob if it fulfils two requirements: 1) the KLD distance is smaller than threshold $\phi$, and 2) the intersection over union of $b$ and $\tilde{b}$ is above threshold $\delta$. This set of perturbations is repeated for $k$ iterations with increasing perturbation intensity, in which random affine transformations sample from increasing ranges of rotations and translations. The KLD distance ensures that color intensity distribution of pixels in blobs are similar to each other, while the intersection over union threshold requires that we expand gradually, without creating disjoint blobs. The increased perturbations also ensure that we first explore neighboring regions to obtain successive expansion. 

\begin{figure}[t!]
\setlength\tabcolsep{1pt}
\def\arraystretch{0.5}
\centering
\begin{tabular}{cccc}
% \begin{tabularx}{\textwidth}{ccccccc}
    % Order: RGB, MNDWI, GT, Threshold, Distance, Refined Mask
    \includegraphics[width=0.24\linewidth]{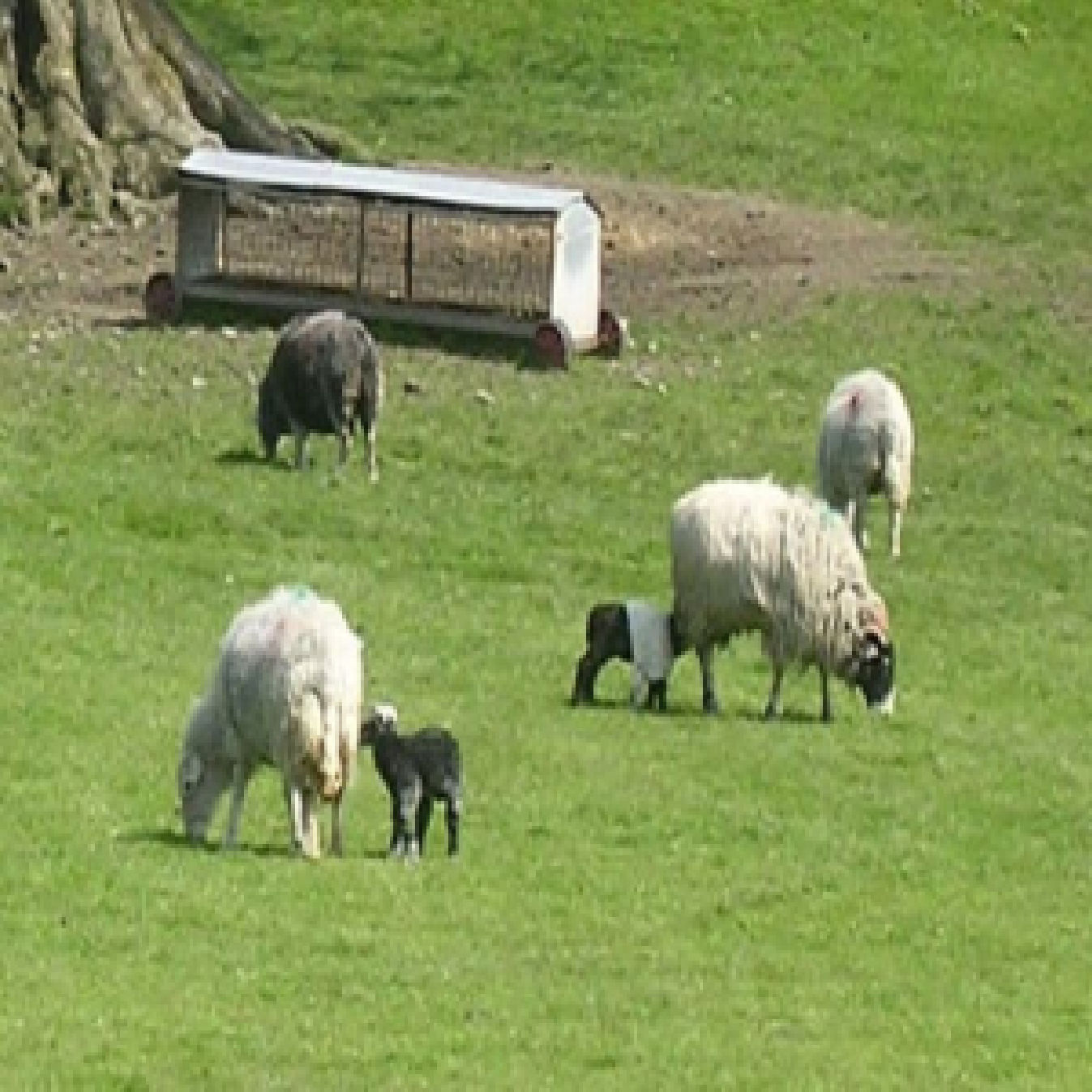}&
    \includegraphics[width=0.24\linewidth]{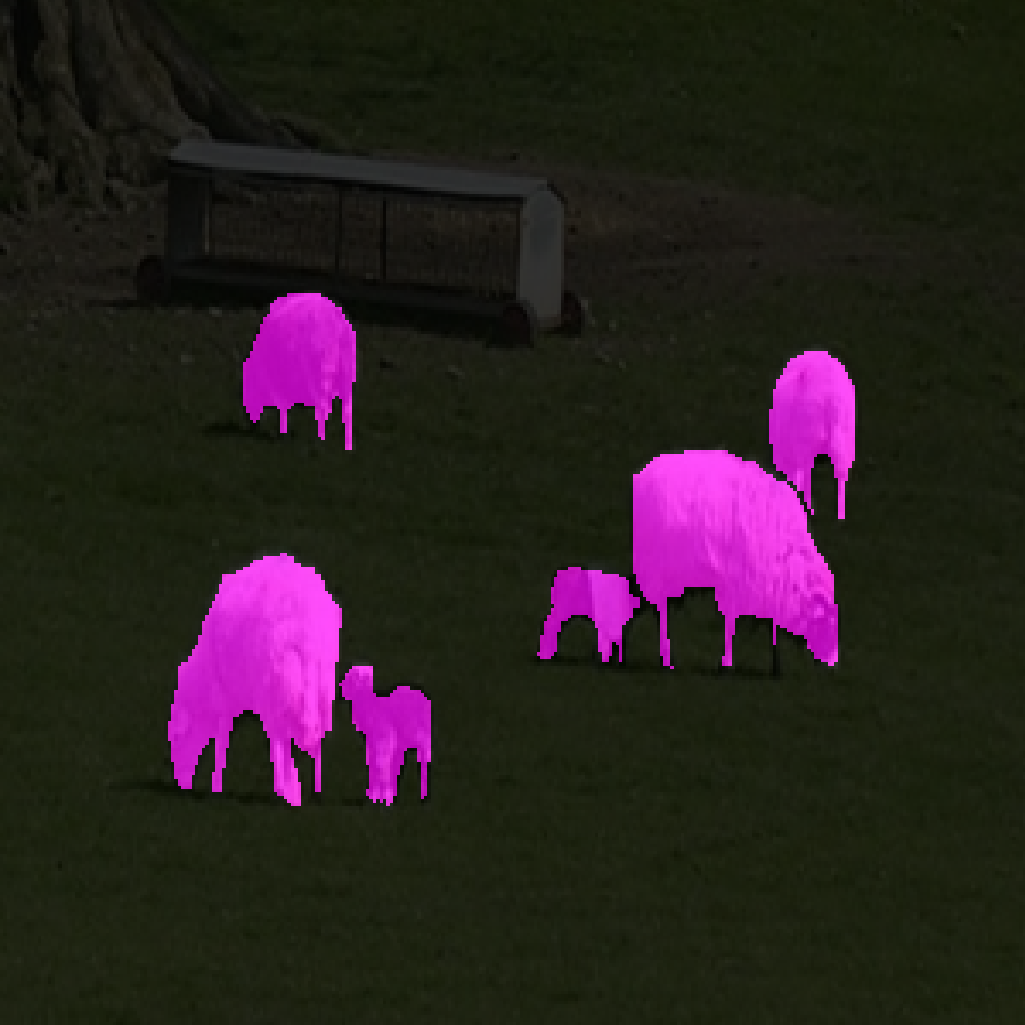} & 
    \includegraphics[width=0.24\linewidth]{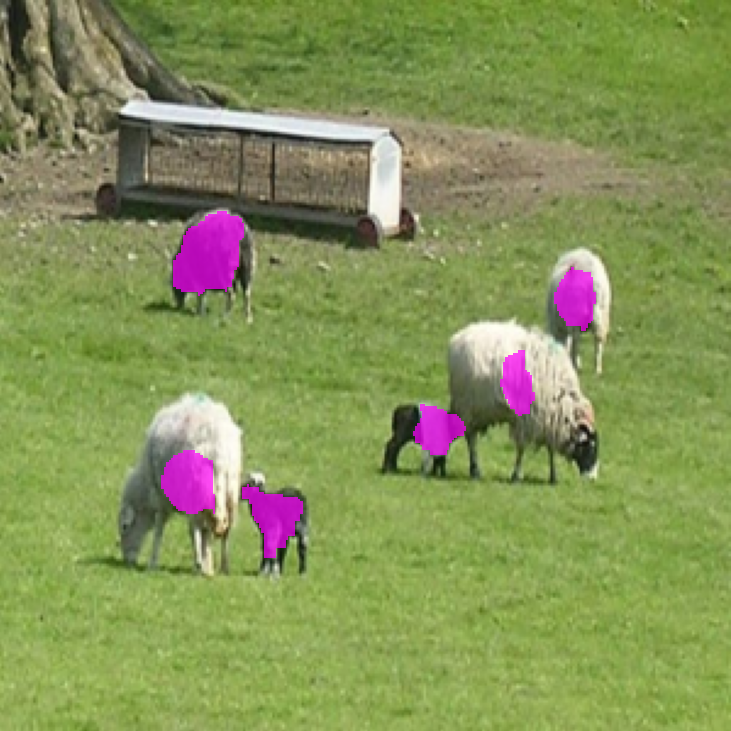}&
    \includegraphics[width=0.24\linewidth]{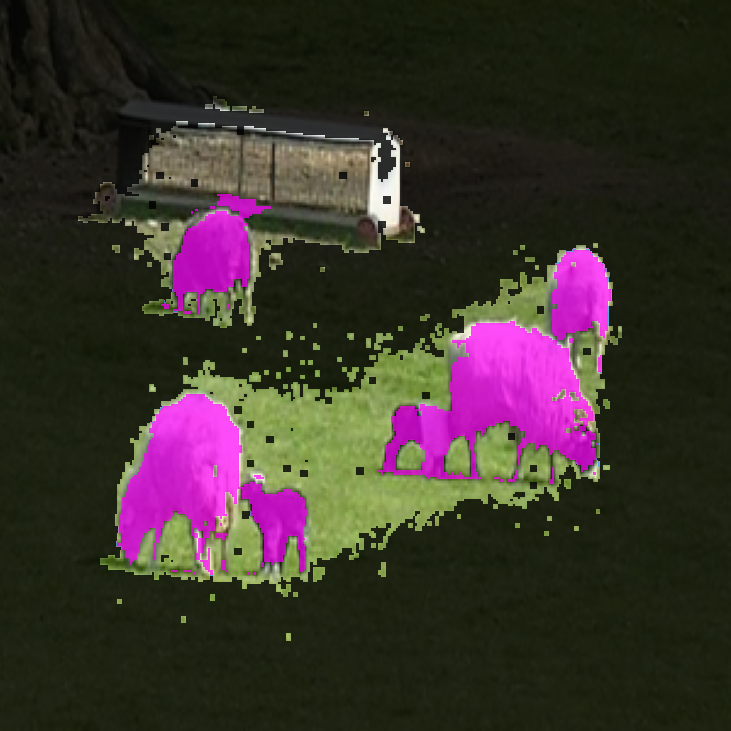} \\
   \includegraphics[width=0.24\linewidth]{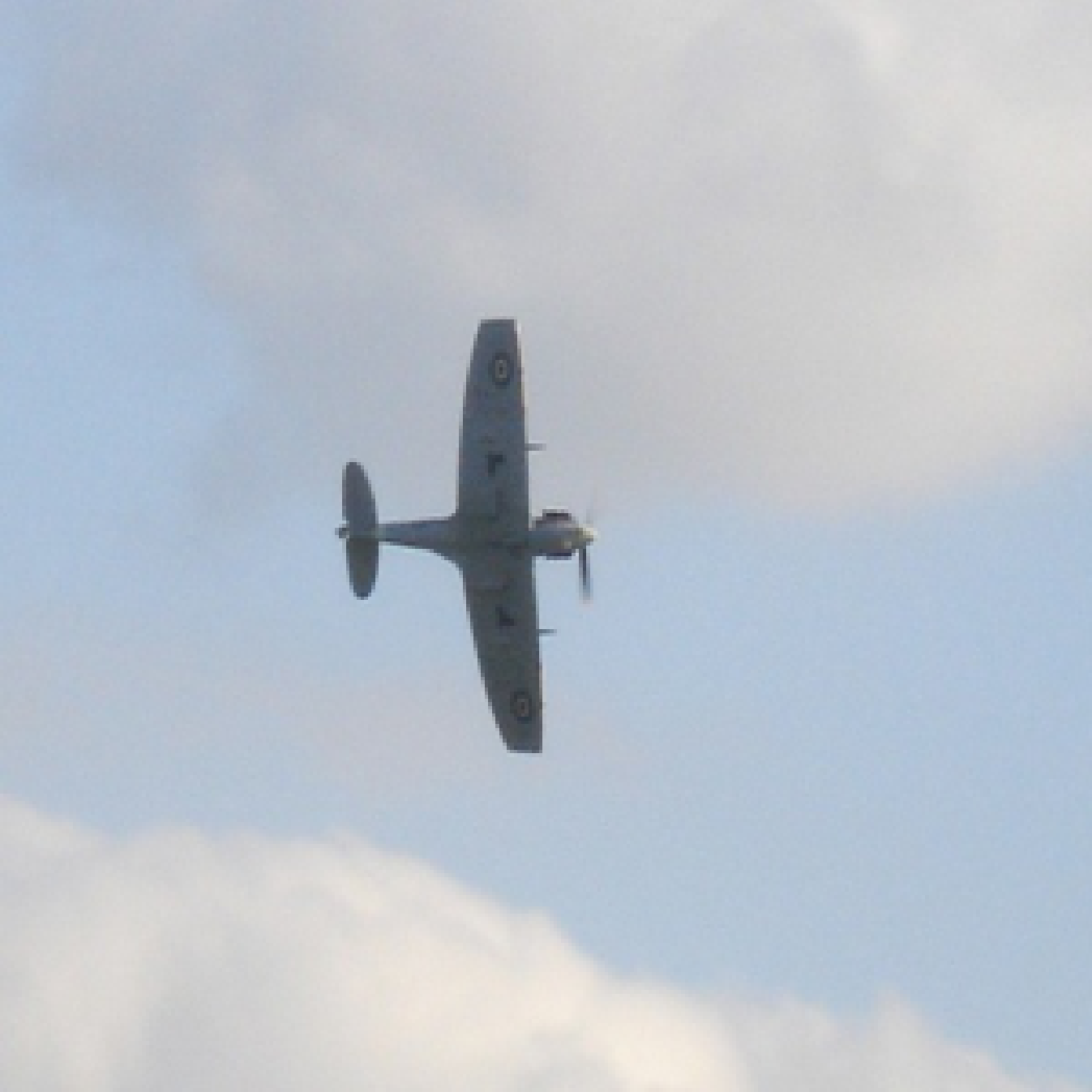}&
    \includegraphics[width=0.24\linewidth]{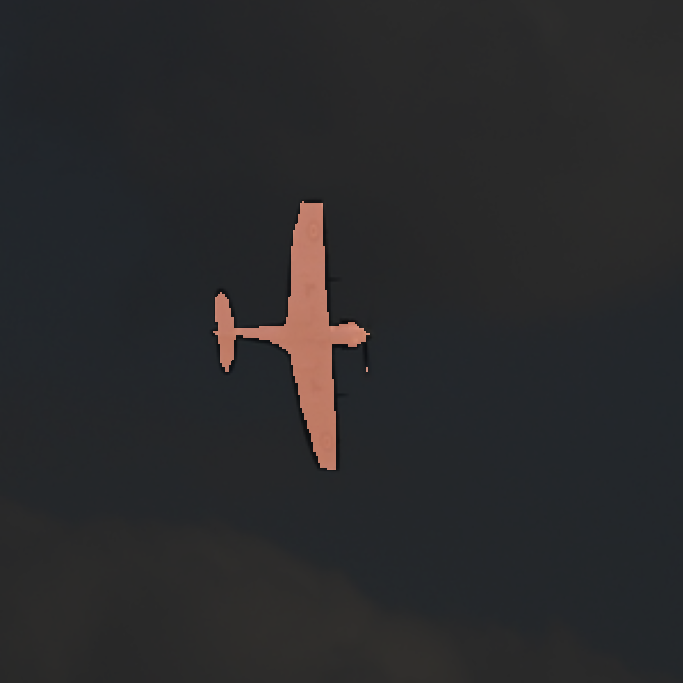} & 
    \includegraphics[width=0.24\linewidth]{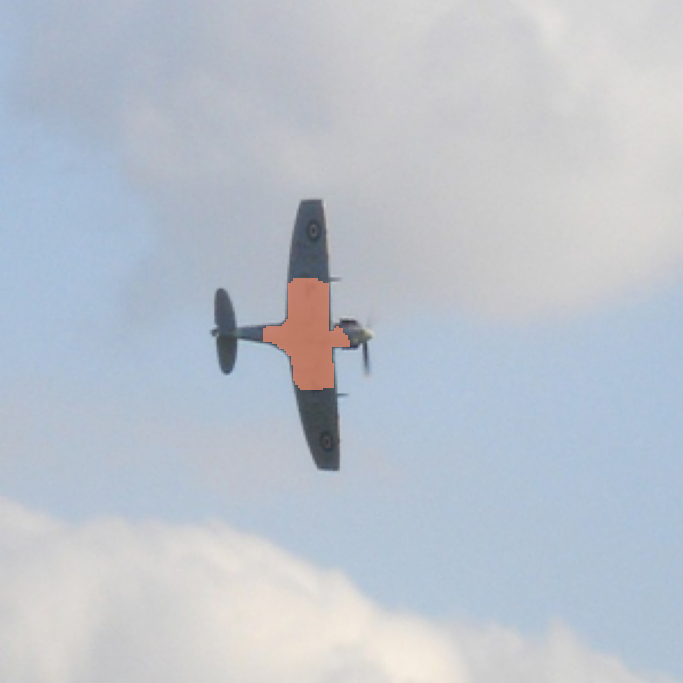} &  
    \includegraphics[width=0.24\linewidth]{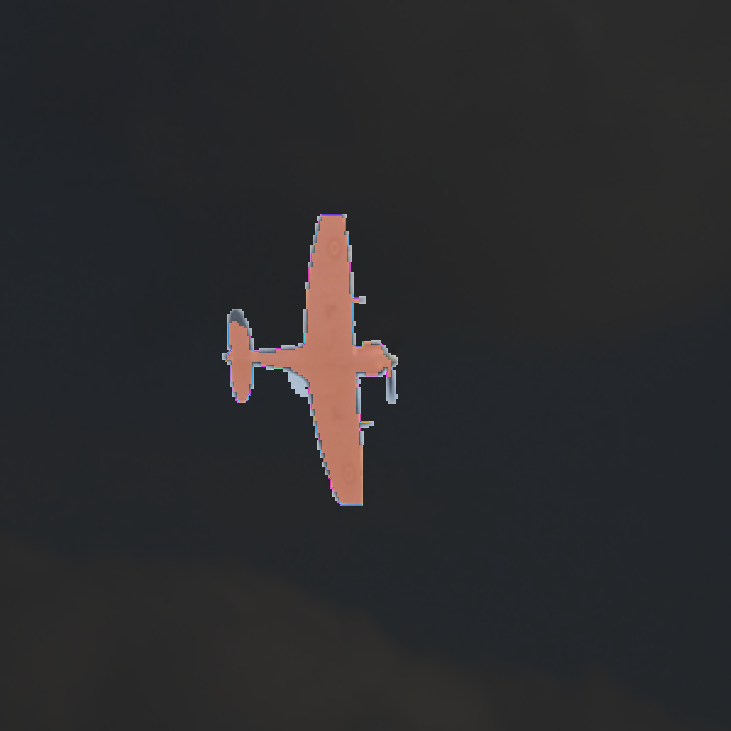} \\
      \includegraphics[width=0.24\linewidth]{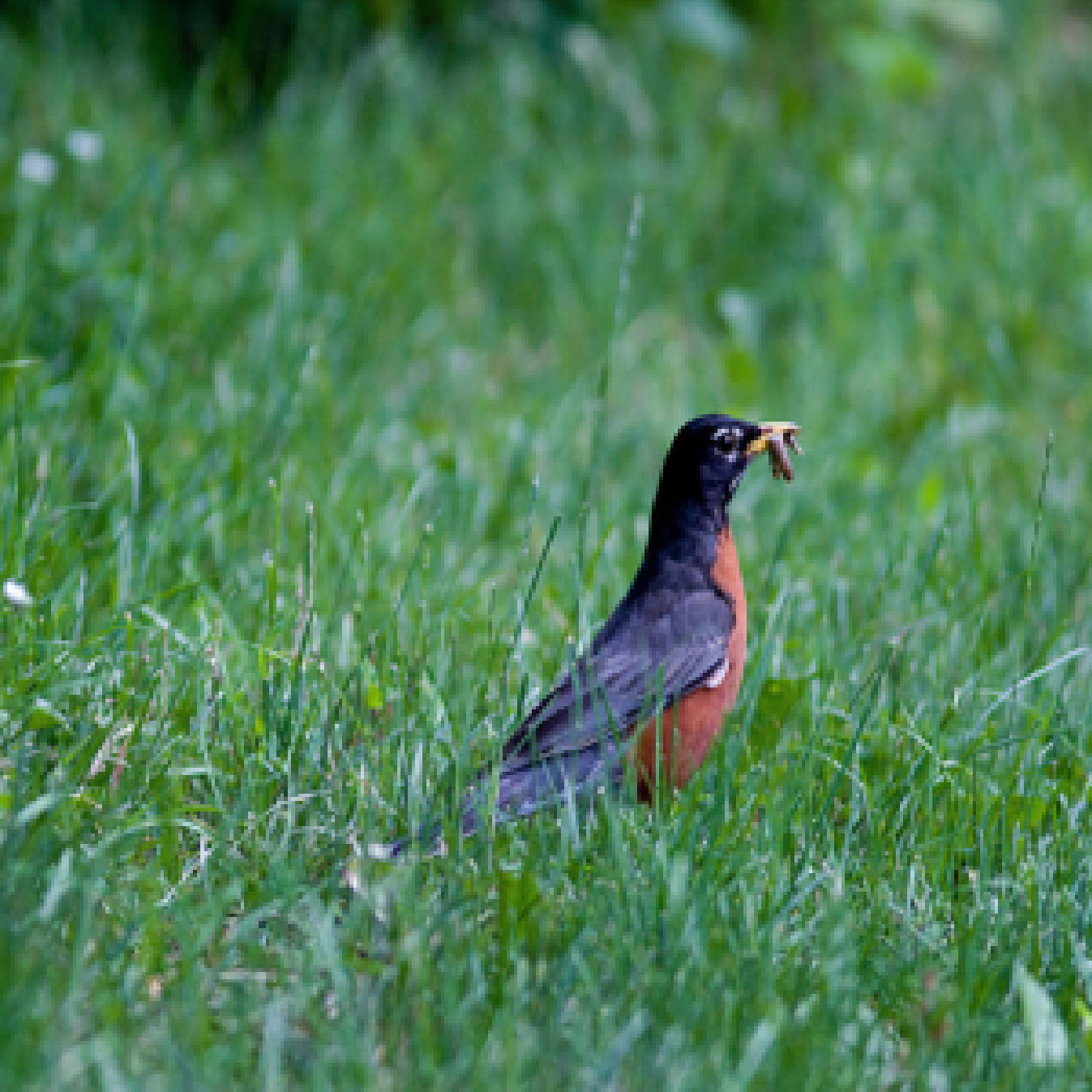}&
    \includegraphics[width=0.24\linewidth]{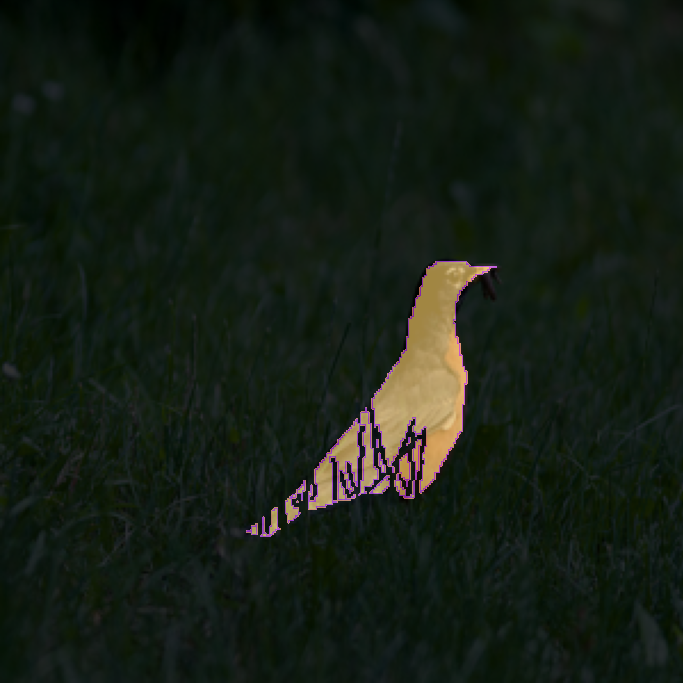} & 
    \includegraphics[width=0.24\linewidth]{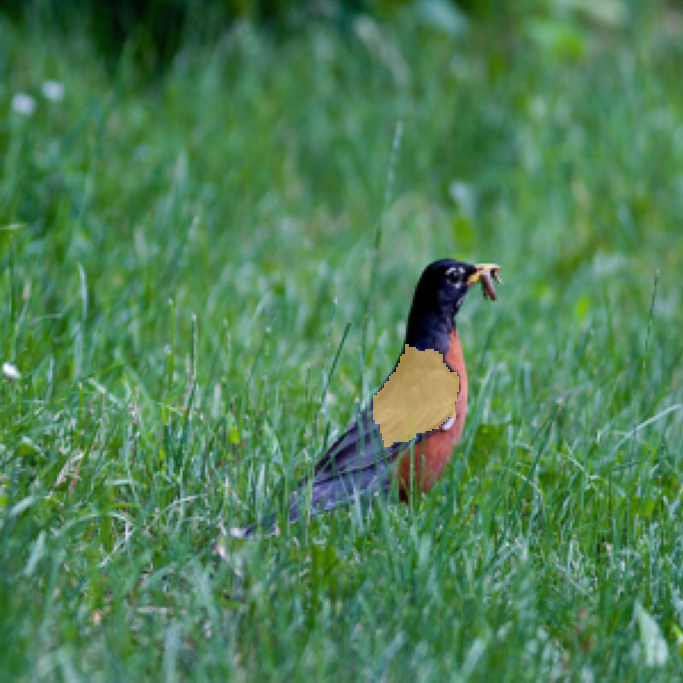} &  
    \includegraphics[width=0.24\linewidth]{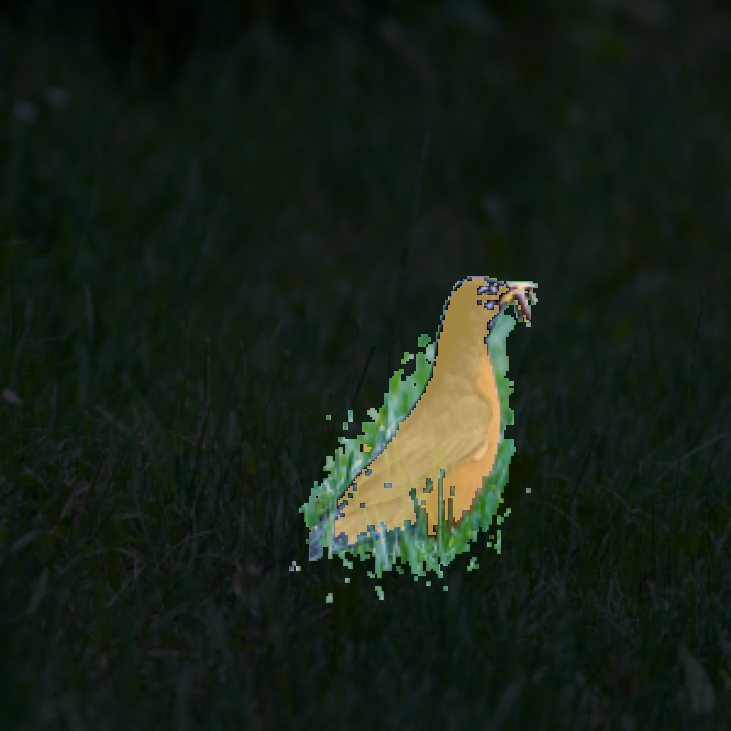} \\
     \small Input Image & 
    \small Ground Truth & 
    \small Point Blot &
     \small Pseudo-Mask 
\end{tabular}
% \end{tabularx}
\vspace{-0.75em}
\caption{\textbf{Qualitative results} of our generated pseudo-masks and point blots on Pascal VOC \cite{Everingham10} training set. Our method provides pseudo-masks converging towards fully-supervised ground truth, allowing for better performance, even when training from scratch. Pixels with low certainty (not color coded) are ignored. Dark gray pixels represent background class. Best viewed in color and zoomed.}
\vspace{-2.em}
\label{fig:relative_entropy_expan}
\end{figure}

The point blot generation pipeline can be seen in Figure \ref{fig:REPE}, and output samples in Figure \ref{fig:relative_entropy_expan}. The method allows us to capture additional neighboring pixels around points without sacrificing excessive computation resources, increasing computational time per iteration by roughly 9.42\%. 

\vspace{-0.9em}
\section{Experiments}
\vspace{-0.4em}
\label{sec:experiments}
We train and evaluate the performance of our method on four datasets: Pascal VOC 2012 \cite{Everingham10}, Cranberry from Aerial Imagery Dataset (CRAID)  \cite{akiva2020finding}, CityPersons  \cite{Shanshan2017CVPR, Cordts2016Cityscapes}, and Inria Aerial Dataset (IAD) \cite{maggiori2017can}. The first is to illustrate our method's performance on a standard benchmark dataset, and the rest are examples of real world applications.
While standard benchmark datasets are essential for baseline efficacy assessment,
we want to demonstrate our method's generalizability and flexibility in domains more common in real world applications.
% real world applications are typically evaluated and structured to maximize efficiency and performance, often formulated as a set of binary tasks. 
% 
We select the CRAID, CityPersons, and IAD datasets because they provide point annotations (upfront or through a pre-processing step) for objects and backgrounds and fully supervised evaluation sets.
Pascal VOC 2012 dataset contains 12,031 images, with 10,582 used for training, and 1,449 for validation. We obtain points for objects by selecting the center points of bounding boxes, and points for backgrounds by uniformly sampling four points per object outside of all boxes in a given scene. Since the center points of objects may not always be located on the object, creating bias in our ground truth, we believe that our method will work even better with datasets that were annotated using points. 
\begin{table}[t!]
\centering
\resizebox{0.48\textwidth}{!}{%
% \begin{tabular}{@{}c{\hspace{2em}}ccc@{\hspace{2em}}}
\begin{tabular}{lccc}
\toprule
Dataset & \multicolumn{3}{c}{Pascal VOC 2012 \cite{Everingham10}} \\ \midrule
Method & Supervision & val & test \\ \midrule
\multicolumn{4}{l}{\centering \footnotesize \textit{Single Stage, Full Supervision }} \\ \midrule
WideResNet38  \cite{wu2019wider} & $\mathcal{F}$ & 80.8 & 82.5 \\
% U-Net \cite{ronneberger2015u} & $\mathcal{F}$ & 80.9 & 83.2   \\
DeepLab v3 \cite{chen2017deeplab}  & $\mathcal{F}$ & - & 87.8  \\
\midrule
\multicolumn{4}{l}{\centering \footnotesize \textit{Multi Stage + Saliency}} \\
\midrule
Bearman \etal \cite{bearman2016s} & $\mathcal{S, P}$ & 46.0 & 43.6 \\
MDC \cite{wei2018revisiting} & $\mathcal{S, I}$ & 60.4 & 60.8 \\
MCOF \cite{wang2018weakly} & $\mathcal{S, I}$ & 60.3 & 61.2 \\
% DCSP \cite{chaudhry2017discovering} & $\mathcal{S, I}$ & 60.8 & 61.9 \\ 
% SeeNet \cite{hou2018self} & $\mathcal{S, I}$ & 63.1 & 62.8 \\
DSRG \cite{huang2018weakly} & $\mathcal{S, I}$ & 61.4 & 63.2 \\
BoxSup \cite{dai2015boxsup} & $\mathcal{S, B}$ & 62.0 & 64.6\\
CIAN \cite{fan2020cian} & $\mathcal{S, I}$ & 64.1 & 64.7\\
FickleNet \cite{lee2019ficklenet} & $\mathcal{S}$ & 64.9 & 65.3\\

\midrule
\multicolumn{4}{l}{\centering \footnotesize \textit{Multi Stage}} \\
\midrule
AffinityNet \cite{ahn2018learning}  & $\mathcal{I}$ & 61.7 & 63.7   \\
IRNet \cite{ahn2019weakly} & $\mathcal{I}$ & 63.5 & 64.8\\
SSDD \cite{shimoda2019self} & $\mathcal{I}$ & 64.9 & 65.5\\
SEAM  \cite{wang2020self} & $\mathcal{I}$ & 64.5 & 65.7   \\
\midrule
\multicolumn{4}{l}{\centering \footnotesize \textit{Single Stage}} \\
\midrule
EM \cite{papandreou2015weakly}  & $\mathcal{I}$ & 38.2 & 39.6  \\
MIL-LSE \cite{pinheiro2015image} & $\mathcal{I}$ & 42.0 & 40.6\\
CRF-RNN \cite{roy2017combining} & $\mathcal{I}$ & 52.8 & 53.7\\
Araslanov \etal \cite{araslanov2020single} & $\mathcal{I}$ & 59.7 & 60.5\\
Araslanov \etal + CRF \cite{araslanov2020single} & $\mathcal{I}$ & 62.7 & 64.3\\
\midrule
Ours & $\mathcal{P}$ &  60.7 &  60.8 \\
Ours + CRF & $\mathcal{P}$ &  62.9 &  63.8 \\
% Ours (w/ pre-trained backbone) & $\mathcal{P}$ &  \textbf{-} &  \textbf{-} \\
\bottomrule
\end{tabular}%
}
\vspace{-0.8em}
\caption{mIoU (\%) accuracy (higher is better) on Pascal VOC 2012 validation and test sets \cite{Everingham10}. $\mathcal{F}$, $\mathcal{I}$, $\mathcal{B}$, $\mathcal{S}$, and $\mathcal{P}$ represent full, image, box, saliency, and point level annotations respectively. Our method achieves SOTA performance even with from-scratch training. Class-wise performance is available in supplementary material.}
\vspace{-1.7em}
\label{tab:results}
\end{table}
For our real world examples, CRAID \cite{akiva2020finding}, a computational agriculture dataset, provides 1022 images with point annotations, and 231 with pixel-wise annotations, CityPersons \cite{Shanshan2017CVPR}, a pedestrian detection dataset subset of Cityscapes \cite{Cordts2016Cityscapes}, provides 2115 training and 391 testing image with bounding boxes (processed to points similar to Pascal VOC), and IAD \cite{maggiori2017can}, a remote sensing dataset, provides 180 images (cropped to 29239 images) with pixel-wise annotations (processed to points).

\vspace{-0.5em}
\subsection{Implementation Details}
\vspace{-0.1cm}
To highlight the contribution of our method, we choose to adopt a standard fully convolutional network (\textbf{\textit{untrained}} ResNet50 backbone encoder) that is trained from scratch. Note that this is not typical of other baseline methods, in which pre-trained, complex networks (often pre-trained on the benchmark or similar dataset) are used to achieve SOTA performance. Our network is trained using the SGD optimizer, with starting learning rate of 1e-5 and cosine annealing scheduler \cite{loshchilov2016sgdr}. 
% Instead of the traditionally used batch normalization layers \cite{ioffe2015batch}, 
We use weight standardization \cite{wu2018group} and group normalization layers \cite{weightstandardization} with group size of 32. Training data is augmented with normalization transformation, color jittering, and random vertical and horizontal flips. We use Cross Entropy loss for training, with ``0" labels ignored (background points, labeled as $\mathcal{C}+1$, are considered instead). For the PAC Refinement Network, we use 12 layers with kernel sizes $(7,7,5,5,3,3,3,3,3,3,3,3)$, dilations $(1,1,2,2,4,4,8,8,16,16,32,32)$, and strides $(2,2,2,2,1,1,1,1,1,1,1,1)$. We use $-0.025, 0.025$ for lower and upper limits for the Expanding Distance Fields, and 0.75 for pseudo-mask thresholding.
For performance evaluation, we report mean Intersection over Union (mIoU) for both validation and test sets. 
Note that all experiments reported in the main paper are done in a single stage,
%The network is trained at a single phase,
without pruning or eliminating output predictions. Baseline method for real world datasets was trained in accordance with the method's reported procedure. 
% We find mIoU using
% \begin{equation}
%     \begin{aligned}
%         &\text{mIoU} = \frac{1}{n} \sum_{i=1}^{n} \frac{y_i \cap \Tilde{y}_i}{y_i \cup \Tilde{y}_i} \text{ ,}
%     \end{aligned}
%     \label{metrics}
% \end{equation}
% where $y_i$, $\Tilde{y}_i$, and $n$ represent the ground truth mask, predicted mask, and number of examples respectively. 
% 
% This work reports on two experiments: single stage, and multi stage pipelines. In the single stage approach, we simply train the network and use the pseudo masks as proposed in the method, without pruning or eliminating output predictions. In the multi stage experiment, we train the network, and then evaluate on the same dataset, pruning output channels corresponding to classes not present in the image. We then save those predictions, and train a separate semantic segmentation network as a fully supervised method. The single stage results are reported in table \ref{tab:results}, and the multi stage results are reported in table \ref{tab:multi_stage_results}.
% 
% Our real world single-stage baseline was trained with accordance to instructions provided in method.
% We also provide our method's performance with pruned outputs (a practice common in multi-stage approach) in the supplementary material. 

\begin{table}[t!]
\centering
\resizebox{0.48\textwidth}{!}{%
% \begin{tabular}{@{}c{\hspace{2em}}ccc@{\hspace{2em}}}
\begin{tabular}{lcccc}
\toprule
Dataset & & CRAID \cite{akiva2020finding} & CityPersons \cite{Shanshan2017CVPR} & IAD \cite{maggiori2017can}\\ \midrule
Method & Sup. & \multicolumn{3}{c}{mIoU (\%)}\\ \midrule
% \multicolumn{4}{l}{\centering \footnotesize \textit{Single Stage}} \\
% \midrule
DeepLab v3 \cite{chen2017deeplab} & $\mathcal{F}$ & 81.3 & 80.7 & - \\ 
ICT-Net \cite{chatterjee2019semantic} & $\mathcal{F}$ & - & - & 80.3 \\
\midrule
Araslanov \etal \cite{araslanov2020single} & $\mathcal{I}$  & 54.9 & 48.2 & 57.6 \\
% Araslanov \etal + CRF \cite{araslanov2020single} & $\mathcal{I}$  & 56.1& 48.7 & 57.9\\
Triple-S \cite{akiva2020finding} & $\mathcal{P, D}$  &  68.7 & - & - \\
\midrule
Ours & $\mathcal{P}$ &  \textbf{72.1} & \textbf{62.8} & \textbf{72.4} \\
% Ours (w/ pre-trained backbone) & $\mathcal{P}$  &  \textbf{72.6} \\
\bottomrule
\end{tabular}%
}
\vspace{-0.8 em}
\caption{mIoU (\%) accuracy on CRAID \cite{akiva2020finding}, CityPersons \cite{Shanshan2017CVPR}, and IAD \cite{maggiori2017can} test sets. Our method is generalizable to arbitrary datasets, significantly outperforming our single-stage baselines on the selected real world datasets.}
\vspace{-2.em}
\label{tab:craid_results}
\end{table}

% \subsection{Baselines}

% 
\vspace{-0.8em}
\section{Results}
\vspace{-0.6em}
\label{sec:results}
% \subsection{Quantitative}
Table \ref{tab:results} presents comparisons between SOTA baselines and our method on the Pascal VOC dataset. In the single-stage approach, our base method outperforms \cite{araslanov2020single} by 1\% on validation and 0.3\% on test sets, even though we train our network from scratch and \cite{araslanov2020single} uses a pre-trained backbone. It is important to note that, as shown in \cite{zhou2016learning, selvaraju2017grad, Oquab_2015_CVPR}, localization is ``free" when a trained classification network (i.e.\ pre-trained backbone) is available, which serves a similar role to our ground truth points. From our experiments, without the usage of a pre-trained backbone, \cite{araslanov2020single} performs significantly worse. On the other hand, using points removes the necessity of separately train a classification network or use pre-trained weights, allowing our method to be used in a broader context and on non-standard datasets without incurring significant additional annotation costs. Table \ref{tab:craid_results} demonstrates the wider range of the capabilities of our method, which performs significantly better than the single-stage baseline (\cite{araslanov2020single}) on the CRAID, CityPersons, and IAD datasets. The poor performance of \cite{araslanov2020single}, or other image-level label driven methods, stems from the dependency on the preceding classification task to provide good class activation maps essential for localization. 
\begin{table}[t!]
\centering
\resizebox{0.47\textwidth}{!}{%
% \begin{tabular}{@{}cccc@{\hspace{1em}}}
\begin{tabular}{@{}c@{\hspace{0.68em}}c@{\hspace{0.68em}}c@{\hspace{0.68em}}c@{}}
\toprule
Expanding Distance Fields & Point Blot & PAC Refiner & mIoU (\%)\\ \midrule
% \multicolumn{4}{l}{\centering \footnotesize \textit{Single Stage}} \\
% \midrule
 \multicolumn{3}{c}{\textit{points only}} & 15.2 \\ 
 & $\checkmark$ & $\checkmark$  & 24.7 \\ 
 & $\checkmark$ &   & 49.1 \\ 
$\checkmark$ &   &  & 38.3 \\ 
$\checkmark$ & $\checkmark$ &   & 48.9 \\ 
$\checkmark$ &   & $\checkmark$ & 54.5 \\ 
$\checkmark$ & $\checkmark$ & $\checkmark$ &  \textbf{60.7} \\ 
% Ours (w/ pre-trained backbone) & $\mathcal{P}$  &  \textbf{72.6} \\
\bottomrule
\end{tabular}%
}
\vspace{-0.8 em}
\caption{\textbf{Ablation study} on Pascal VOC 2012 validation set \cite{Everingham10}. We investigate the effects of each component in our proposed method and show its impact on overall performance. We consider five variations of our modules with their respective performance shown above. Note when point blots are not used, points are used instead. If PAC Refiner or Expanding Distance Fields is used, then pseudo-mask is generated from thresholded output (refined or not) features. It can be seen from the result that thresholding refined features alone is not enough, and that spatially accurate features obtained by the expanding distance fields are essential in generation of better pseudo-masks and performance.}
\vspace{-2.2em}
\label{tab:ablation}
\end{table}
When the images have few (or binary) classes, the classification becomes too easy which results in less refined feature outputs and sub-optimal localization and CAM coverage. The effects of such dependencies are greatly magnified with respect to the number of objects in the scene, making scenes such as in CRAID and CityPersons, which have large counts of small objects, increasingly difficult for any method using image-level labels for segmentation. This can also be observed in the qualitative results, with our baseline producing coarse outputs for CRAID and CityPersons datasets. In contrary, images in Pascal VOC 2012 have an average of 2.37 objects per image, making it easier to generate features without spatial guidance. Thorough empirical analysis of performance degradation with decrease in object sizes and increase in counts is explained in \cite{zhou2018weakly}. Additional class-wise quantitative results of our method and the baselines for Pascal VOC 2012 and qualitative results for CRAID, CityPersons, and IAD datasets are available in the supplementary material.
% 
% We also outperform all other single stage methods by a considerable margin. Our single-stage approach also outperforms all multi-stage methods
% 
% \subsection{Ablation Study}
% For ablation study, we investigate the effects of each component in our proposed method and show its impact on overall performance. We consider five cases detailed in table \ref{tab:ablation}. 

\vspace{-0.2em}
\subsection{Conclusion}
\vspace{-0.5em}
This paper presents a practical single-stage weakly supervised semantic segmentation method applicable to non-standard datasets for which pre-trained backbones are not available, or pre-training classification task is insufficient. By utilizing our expanding distance fields and point blots, our method is able to achieve SOTA performance on the benchmark dataset as well as significantly better performance than SOTA methods on real-world domains.

\newpage

{\small
\bibliographystyle{ieee_fullname}
\bibliography{pmp}
}
% egbib
\end{document}